\documentclass[unnumsec,webpdf,contemporary,large]{oup-authoring-template}%
\usepackage{booktabs}
\graphicspath{{Fig/}}

\usepackage{pifont}
\usepackage{makecell}
\usepackage[table]{xcolor}
\newcommand{\cmark}{\ding{51}}
\newcommand{\xmark}{\ding{55}}
\usepackage{comment}

\theoremstyle{thmstyleone}%
\theoremstyle{thmstyletwo}%
\theoremstyle{thmstylethree}%

\begin{document}

\journaltitle{}
\DOI{DOI added during production}
\copyrightyear{YEAR}
\pubyear{YEAR}
\vol{XX}
\issue{x}
\access{Published: Date added during production}
\appnotes{Problem Solving Protocol}

\firstpage{1}
\begin{comment}
\journaltitle{Journal Title Here}
\DOI{DOI added during production}
\copyrightyear{YEAR}
\pubyear{YEAR}
\vol{XX}
\issue{x}
\access{Published: Date added during production}
\appnotes{Paper}

\firstpage{1}

%\subtitle{Subject Section}

\title[Short Article Title]{Article Title}

\author[1,$\ast$]{First Author}
\author[2]{Second Author}
\author[3]{Third Author}
\author[3]{Fourth Author}
\author[4]{Fifth Author\ORCID{0000-0000-0000-0000}}

\address[1]{\orgdiv{Department}, \orgname{Organization}, \orgaddress{\street{Street}, \postcode{Postcode}, \state{State}, \country{Country}}}
\address[2]{\orgdiv{Department}, \orgname{Organization}, \orgaddress{\street{Street}, \postcode{Postcode}, \state{State}, \country{Country}}}
\address[3]{\orgdiv{Department}, \orgname{Organization}, \orgaddress{\street{Street}, \postcode{Postcode}, \state{State}, \country{Country}}}
\address[4]{\orgdiv{Department}, \orgname{Organization}, \orgaddress{\street{Street}, \postcode{Postcode}, \state{State}, \country{Country}}}

\corresp[$\ast$]{Corresponding author. \href{email:email-id.com}{email-id.com}}
\end{comment}
\title[GenoMorph: Pathway-Grounded Genomic Disease Reasoning]{{\normalfont\bfseries\itshape GenoMorph}: Pathway-Grounded Genomic Disease Reasoning via Adaptive Latent Computation}

\author[1,2,\dag]{Tanmoy Kanti Halder}
\author[1,\dag]{Akash Ghosh}
\author[1]{Arijit Roy}
\author[1,$\ast$]{Sriparna Saha}
\corresp[$\ast$]{Corresponding author. \href{mailto:sriparna@iitp.ac.in}{sriparna@iitp.ac.in} (Sriparna Saha)}

\address[1]{\orgdiv{Department of Computer Science and Engineering}, \orgname{Indian Institute of Technology Patna}, \orgaddress{\street{Bihta}, \postcode{801106}, \state{Bihar}, \country{India}}}
\address[2]{\orgname{Prasannadeb Women's College}, \orgaddress{\country{India}}}

\received{Date}{0}{Year}
\revised{Date}{0}{Year}
\accepted{Date}{0}{Year}

%\editor{Associate Editor: Name}

%\abstract{
%\textbf{Motivation:} .\\
%\textbf{Results:} .\\
%\textbf{Availability:} .\\
%\textbf{Contact:} \href{name@email.com}{name@email.com}\\
%\textbf{Supplementary information:} Supplementary data are available at \textit{Journal Name}
%online.}

\abstract{Large language models (LLMs) have demonstrated strong capabilities in biological reasoning; however, genomic disease inference remains largely dependent on memorized gene-disease associations rather than understanding of biological pathways.This shortcut learning undermines robustness and generalization, and it breaks down entirely when explicit molecular identifiers are unavailable. To overcome these limitations , we present {\normalfont\bfseries\itshape GenoMorph}, a multimodal genomic reasoning framework that shifts disease prediction from associative gene–disease mapping toward pathway-grounded  reasoning. {\normalfont\bfseries\itshape GenoMorph} couples a frozen DNA foundation model with question-conditioned cross-attention fusion, self-adaptive latent reasoning (LatentSp), a residual reasoning gate for iterative genomic evidence reinjection, and rejection sampling fine-tuning regularized by hierarchical optimal transport (OT). Instead of learning direct mappings between genes and diseases, the proposed framework progressively aligns genomic sequence representations with latent pathway dynamics, enabling reasoning trajectories that follow underlying molecular interactions before producing disease predictions. Furthermore, the self-adaptive latent reasoning mechanism dynamically allocates computation according to reasoning confidence, reducing unnecessary latent reasoning steps and substantially improving inference efficiency. To understand the impact of our framework, we build an anonymized  benchmark from Kyoto Encyclopedia of Genes and Genomes(KEGG) pathway that swaps every gene and molecular identifier for globally consistent anonymous symbols while preserving sequences and pathway topology, removing memorization shortcuts. {\normalfont\bfseries\itshape GenoMorph} raises the weighted F1 from 0.7863 (BioReason) to 0.9412, and rejection sampling fine-tuning with self-adaptive latent reasoning pushes it to 0.9725 while cutting latency nearly 60\%. On the anonymized benchmark it reaches 0.9465 F1, substantially outperforming prior systems and confirming that accurate disease prediction can arise from pathway reasoning rather than memorized gene–disease associations.
}

% \otherabstract[Additional Abstract]{Use this element for elements such as Graphical abstract, Lay summary, Translated abstract etc. Que cum aut etum qui ium dolupta ssequia autati odis demporepe ad et es alit rem repudaerae min et volorum re volupta nobit volectur aut fuga.}

\otherabstract[Graphical Abstract]{
\begin{center}
\includegraphics[width=0.85\textwidth]{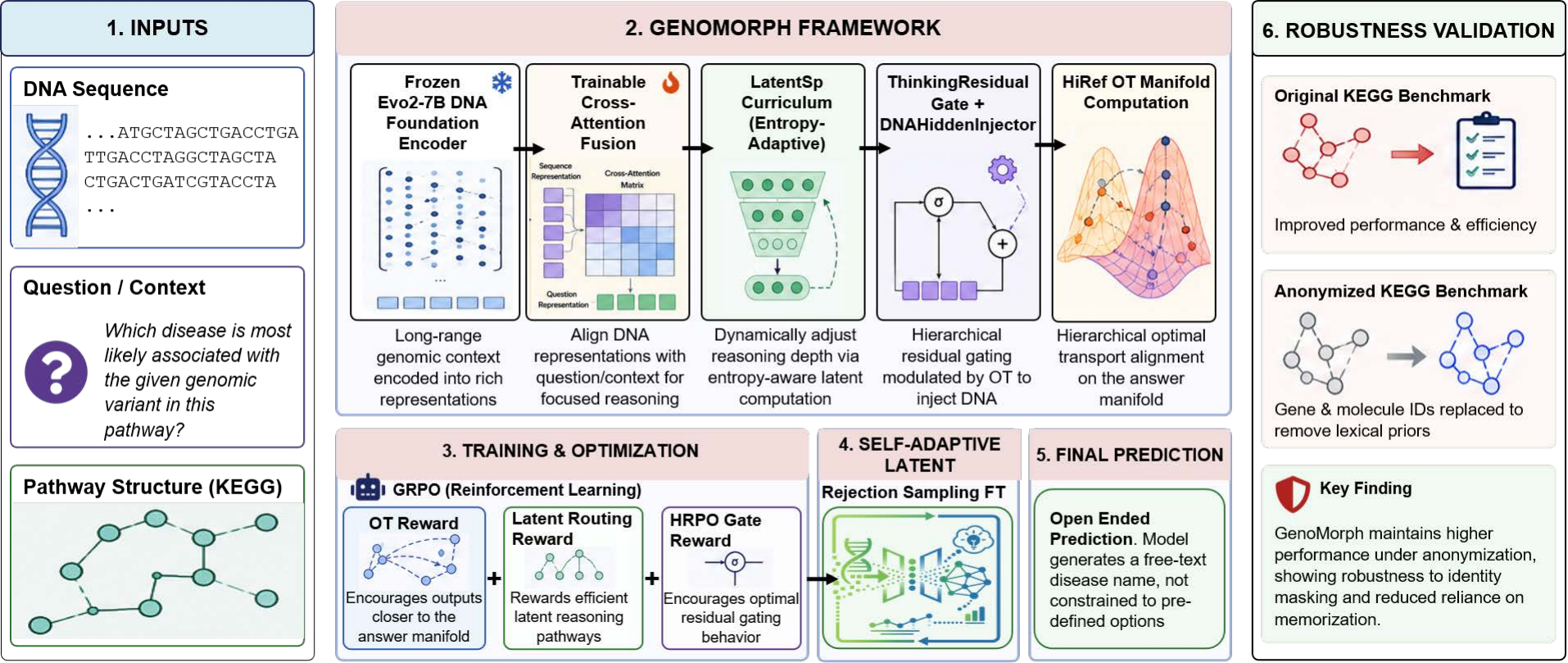}
\end{center}

\vspace{0.5em}

{\small
\noindent Overview of {\normalfont\bfseries\itshape GenoMorph}, a multimodal genomic reasoning framework for open-ended disease prediction. The model integrates DNA sequence and biomedical context through cross-attention fusion, entropy-adaptive latent reasoning, and residual biological injection, with training-time alignment to KEGG pathway structure via hierarchical optimal transport, followed by adaptive GRPO optimization for robust disease prediction and finally rejection sampling for self-adaptiveness.
}
}

% \boxedtext{Key Messages}{
% \begin{itemize}
% \item Key boxed text here.
% \item Key boxed text here.
% \item Key boxed text here.
% \end{itemize}}

\maketitle

%\begin{epigraph}
%Epigraph text. Ximporem qui reperov idempedit modio. Bisto imagnatem quae aceptis
%nobitae quid eum rae adignis quias-sit vellacc uptatur sunt quis rentis eaquasit alia deliquam
%rec-to consed unt. Empor sum ratur ressimusdae. Nam fugiae.
%\source{Epigraph source}
%\end{epigraph}

\section{Introduction}

    Recent advances in biological foundation models have substantially expanded the ability of machine learning systems to model genomic, transcriptomic, and regulatory sequences at scale. Large-scale DNA language models, including Evo2~\cite{evo2}, DNABERT-2~\cite{dnabert2}, Nucleotide Transformer~\cite{nucleotide_transformer}, Caduceus~\cite{caduceus}, and Enformer~\cite{enformer} learn contextual representations directly from raw nucleotide sequences and capture rich evolutionary and functional signals that generalize across diverse downstream genomic tasks. These developments provide a powerful foundation for sequence-aware biological intelligence. However, translating genomic representations into reliable disease reasoning remains challenging because biological inference requires jointly reasoning over genomic sequences, molecular interactions, and biological pathway topology rather than learning direct gene–disease associations or sequence statistics alone~\cite{bioreason,MAMMALM}.

    Large language models (LLMs) have recently demonstrated remarkable performance across biological reasoning tasks, including gene-disease association prediction, pathway interpretation, and clinical decision support~\cite{bioreason,medprompt,medpalm2}.However, the reliability of foundation-model outputs remains a critical concern, particularly in high-stakes applications where hallucinated or unsupported predictions can lead to erroneous conclusions and hamper the trustworthiness of these models\cite{sahoo2024comprehensive,ghosh2025clinic,ghosh2026background,xia2024cares}. 
    Hybrid genomic reasoning systems, such as BioReason~\cite{bioreason}, first encode genomic DNA sequences using a DNA foundation model~\cite{evo2} and then project the resulting sequence embeddings into the representation space of a language model to infer diseases from genomic and textual information. Alternative architectures address the same problem differently: GeneGPT~\cite{genegpt} teaches an LLM to call NCBI Web APIs rather than fusing embeddings directly, and TxGemma~\cite{txgemma} offers broad, general-purpose mechanistic reasoning across molecular and clinical properties without ingesting raw DNA. Yet whether embedding-fusion, tool-augmented, or instruction-tuned, these architectures rely on static projection layers, in-context API calls, or supervised fine-tuning, limiting dynamic interaction between genomic evidence and language reasoning. Disease prediction consequently reduces to an associative mapping between gene identities and disease labels rather than reasoning through the pathways that connect sequence variation to phenotype — behavior that limits generalization under unseen genes, aliases, or anonymized entities that participate in the same pathway. This reflects a broader question in biological AI: whether genomic foundation models learn transferable biological abstractions or merely compress sequence statistics~\cite{genome_eval}. \par

Two lines of work suggest a way past this associative ceiling. First, rejection sampling fine-tuning (RSFT) has emerged as an effective paradigm for improving reasoning: methods such as RFT~\cite{yuan2023rft} and RAFT~\cite{raft2023} show that filtering and fine-tuning on high-quality sampled reasoning trajectories, rather than optimizing directly against a learned reward, improves multi-step reasoning, generalization, and efficiency — a natural mechanism for learning biological reasoning policies instead of fitting gene–disease associations. A similar strategy has proven effective in multimodal medical reasoning, where ArogyaSutra\cite{arogya} fine-tunes an actor on successful actor–critic trajectories so that the distilled model no longer requires a critic at inference, a principle we extend to adaptive latent computation in genomic reasoning. Complementary to rejection-based reasoning optimization, recent work has explored curriculum-informed reinforcement learning with Group Relative Policy Optimization (GRPO) for medical reasoning, improving reasoning quality through progressively structured training\cite{onyame2026cure}. Second, reasoning has shifted from explicit tokens toward latent-space computation. Continuous latent reasoning frameworks such as Coconut~\cite{coconut} and its adaptive successors — LatentSeek~\cite{latentseek}, SoftCoT++~\cite{softcotpp}, and the Hybrid Latent Reasoning framework (HRPO)~\cite{hrpo} — allocate additional computation according to prediction uncertainty, echoing the broader finding that scaling inference-time compute by task difficulty can outperform scaling model size~\cite{testtimescaling}. Complementing both directions, geometry-aware optimization improves reasoning fidelity: hierarchical optimal transport methods such as HiRef~\cite{hiref} recursively refine transport plans, while transport-based biological models such as STRAND~\cite{strand} characterize molecular state transitions under perturbation. Together, these developments point toward a common requirement: biologically grounded genomic reasoning needs more than static representation learning, motivating the integration of rejection sampling fine-tuning, adaptive latent computation, and geometry-aware optimization into a single framework. A further gap remains, however. Existing latent-reasoning methods spend a near-identical compute budget on every sample, yet genomic reasoning complexity varies sharply — straightforward pathogenic variants can often be resolved from local molecular evidence, whereas complex disease phenotypes require reasoning across multiple interacting pathways and biological processes. This motivates a self-adaptive latent policy that expands reasoning only when additional biological evidence is required.

Motivated by these observations, we propose {\normalfont\bfseries\itshape GenoMorph}, a multimodal genomic reasoning framework that integrates question-conditioned genomic representation, latent reasoning, geometry-aware optimization, and reinforcement learning.

{\normalfont\bfseries\itshape GenoMorph} is trained through a four-stage pipeline. First, supervised fine-tuning establishes the core reasoning architecture through question-conditioned CrossAttentionFusion~\cite{cross}, curriculum-based~\cite{curriculumlearning} LatentSp training~\cite{latentseek, coconut}, and ThinkingResidualGate ~\cite{hrpo} pretraining for adaptive genomic evidence reinjection. Second, an offline Hierarchical Optimal Transport (HiRef-OT)~\cite{hiref,optimaltransport} procedure measures the geometric discrepancy between genomic and disease-answer representations, producing fixed geometry-aware signals for subsequent optimization. Third, RL methods  Group Relative Policy Optimization (GRPO)~\cite{ppo,grpo,lora,attention} and preference optimization techniques like DPO \cite{rafailov2023direct} showed great results in various downstreaming tasks like improving  disease prediction,  reasoning quality, biological consistency, faithfulness  using a multi-objective reward signal ~\cite{hiref,ghosh2026rado} . Finally, Rejection Sampling Fine-Tuning (RSFT)~\cite{yuan2023rft,raft2023} internalizes the adaptive reasoning behavior learned during GRPO into the model parameters, eliminating the need for explicit online entropy computation during inference. Together, these stages enable {\normalfont\bfseries\itshape GenoMorph} to dynamically allocate latent computation and selectively revisit genomic evidence, promoting pathway-grounded reasoning rather than direct gene--disease associations.

Our contributions can be summarised as follows : 1) We argue that the central weakness of current genomic disease reasoning is not architectural but epistemic where existing DNA–LLM systems predict disease by recognizing gene identities rather than reasoning through the pathways that connect sequence variation to phenotype. We reframe the task as the reconstruction of a biological reasoning trajectory  that shapes learning implicitly, through geometry-aware supervision, rather than being supplied as an explicit lexical shortcut. 2) We propose {\normalfont\bfseries\itshape GenoMorph}, a multimodal genomic reasoning framework that replaces static genomic projection with question-conditioned cross-attention fusion, curriculum-based latent reasoning (LatentSp), and a residual reasoning gate for on-demand genomic evidence reinjection, all regularized by geometry-aware hierarchical optimal transport. A self-adaptive latent policy, distilled through rejection sampling fine-tuning, further allocates computation by reasoning difficulty — expanding it for complex variants and terminating early for simple ones — so that {\normalfont\bfseries\itshape GenoMorph} reasons through pathway dynamics rather than learning direct gene–disease mappings while jointly improving accuracy and inference efficiency. 3) To rigorously test whether disease prediction reflects biological reasoning or mere identity recall, we construct an anonymized KEGG~\cite{kegg} benchmark that replaces every gene and molecular name with globally consistent anonymous symbols while preserving genomic sequences and pathway topology. 4) Our extensive experiments across both the original and anonymized settings show that text-only LLMs collapse once lexical cues are removed, whereas {\normalfont\bfseries\itshape GenoMorph} reasons from sequence and pathway evidence: it lifts the weighted F1 from 0.7863 (BioReason) to 0.9725 while cutting inference latency by nearly 60\%, and retains 0.9465 F1 under anonymization — confirming that its predictions follow the biological pathway connecting genomic variation to phenotype rather than memorized gene–disease associations.

% This is an example of a new paragraph with a numbered footnote\footnote{\url{https://www.academic.oup.com/}} and a second footnote marker.\footnote{Example of footnote text.}
\section{Materials and methods}\label{sec:methods}

In this section, we describe the problem formulation, benchmark construction, and the methodological components of {\normalfont\bfseries\itshape GenoMorph}. We first formulate genomic disease reasoning as a multimodal conditional reasoning task, followed by the construction of the original and anonymized KEGG benchmarks. We then present the {\normalfont\bfseries\itshape GenoMorph} framework, including DNA–language representation alignment, self-adaptive latent reasoning, residual reasoning gate optimization, hierarchical optimal transport reward construction, and rejection sampling fine-tuning. We close the section with the training configuration, evaluation metrics, and baseline models used throughout the Results section.

\subsection{Problem formulation}
Given a genomic variant sequence $S$, a natural-language biomedical query $Q$, and its associated KEGG pathway graph $G=(V,E)$ available only at training time, we seek a predictor $f$ that (i) infers the correct disease $y^\star$ from the multimodal input $X=(S,Q)$, and (ii) does so by reasoning over sequence and pathway evidence rather than memorized gene identities, such that the prediction is preserved when biological entity names are removed. Formally, we require

\begin{equation}
P_{\mathrm{pred}}(f)
=
\left\{
y \;\middle|\;
y = \arg\max_{y} P(y \mid S,Q)
\right\},
\end{equation}

and

\begin{equation}
P_{\mathrm{mech}}(f)
=
\left\{
f(S,Q) \;\middle|\;
f(S,Q) = f\bigl(S,\mathcal{A}(Q)\bigr)
\right\},
\end{equation}

where $\mathcal{A}$ is a global bijection that renames every gene and molecular entity while leaving $S$ and the pathway topology $G$ unchanged. The condition $P_{\mathrm{pred}}$ requires accurate disease inference under standard inputs, while $P_{\mathrm{mech}}$ formalizes that the prediction arises from genomic and pathway evidence---remaining invariant when lexical identifiers are anonymized---rather than from memorized gene--disease associations. Pathway structure $G$ enters only as geometry-aware supervision during training and is never provided at inference.

\subsection{Dataset construction}\label{sec:dataset_construction}
To evaluate genomic disease reasoning, we build our benchmark on the publicly
available KEGG-derived corpus introduced in BioReason. The dataset provides pathway-grounded reasoning instances curated
from biologically validated KEGG disease pathways, pairing curated
variant-disease associations with pathway annotations, and thus supports both
sequence-aware reasoning and pathway-level disease inference while enabling
direct comparison with existing genomic reasoning frameworks.
 
Each instance comprises three components: (i) a DNA sequence spanning a genomic
variant or molecular region, (ii) a natural-language query describing the
reasoning objective, and (iii) the corresponding disease label. Each sample is
additionally linked to its underlying KEGG~\cite{kegg} pathway graph $G=(V,E)$,
which encodes the molecular interactions and regulatory relationships connecting
genomic variation to disease. These graphs are used only for offline
representation alignment and reward construction during training; they are never
provided as inputs at inference.
 
To test whether models rely on lexical biological priors rather than memorized
reasoning, we construct an anonymized version of the benchmark. All explicit
biological identifiers  like gene symbols, protein and enzyme names, KEGG entity
names, and other molecular identifiers are replaced under a globally consistent
scheme in which each entity receives a unique anonymous symbol that remains fixed
across the entire dataset: if \textit{BRCA1} maps to \texttt{gene\_1}, every
occurrence of \textit{BRCA1} in training and evaluation becomes \texttt{gene\_1}.
Crucially, DNA sequences and pathway topology are left unchanged. Formally, for
each pathway graph $G=(V,E)$ we define an anonymization function
\[
G' = \mathcal{A}(G),
\]
where $\mathcal{A}(\cdot)$ is a bijective global relabeling of all biological
entities that preserves graph topology, connectivity, and molecular
interactions. Unlike local masking, this maintains structural consistency across
the benchmark while removing the lexical shortcuts that directly reveal disease
identity. The final benchmark thus provides two complementary settings: (i) the original
KEGG benchmark, which evaluates genomic disease reasoning under standard
conditions, and (ii) the anonymized KEGG benchmark, which isolates mechanistic
robustness by removing explicit biological identities while preserving pathway
structure. Together they let us determine whether disease prediction is driven by
genuine pathway reasoning or by memorized gene-disease associations. Following
the BioReason protocol, all methods and both settings use identical train,
validation, and test splits to ensure fair comparison.

\subsection{Data preprocessing}

Before training, all benchmark samples are transformed into a unified multimodal representation suitable for genomic reasoning. For each instance, the associated DNA sequence is extracted and standardized as the genomic input, while the corresponding natural-language question is retained as the textual reasoning context. Disease labels are normalized into open-ended textual outputs, allowing the model to perform free-form disease generation rather than constrained classification. To improve computational efficiency, genomic representations are precomputed offline using the frozen Evo2-7B~\cite{evo2} foundation model. Specifically, each DNA sequence is encoded once to obtain its hidden representation, which is subsequently cached and reused throughout supervised training and rejection sampling fine-tuning. This preprocessing eliminates repeated forward passes through the frozen DNA encoder, substantially reducing GPU memory consumption and training time.

The textual inputs are tokenized using the Qwen3 tokenizer. Following the BioReason framework, a dedicated placeholder token (\texttt{<|dna\_pad|>}) is inserted into the language input to indicate the position where genomic information will be fused. During training and inference, this placeholder is replaced with the precomputed Evo2 sequence embeddings through the proposed cross-attention fusion module.

To support geometry-aware reinforcement learning, an additional offline preprocessing stage is performed after supervised training. The pretrained cross-attention model is first used to extract fused genomic representations for every training sample. These representations are subsequently aligned with the corresponding disease-answer embeddings using Hierarchical Optimal Transport (HiRef), producing an optimal transport distance for each sample. The resulting transport distances are stored and later incorporated as geometry-aware reward signals during rejection sampling fine-tuning. Since the Monge correspondence and target manifold are computed once and fixed, no transport-plan optimization is required during training or inference; the OT distance to this fixed target is instead recomputed at each training step as the genomic representation evolves.

For the anonymized benchmark, preprocessing is identical except that all biological identifiers are first replaced using the global anonymization mapping described in Section~\ref{sec:dataset_construction}. DNA sequences remain unchanged throughout preprocessing, ensuring that the only biological evidence available to the model is derived from genomic sequence information rather than explicit molecular identities.

\subsection{Preliminaries: Baseline architecture}
{\normalfont\bfseries\itshape GenoMorph} builds directly on BioReason~\cite{bioreason}, a multimodal framework that couples DNA foundation models with large language models for disease prediction and serves as our primary baseline. BioReason encodes genomic sequences with the frozen Evo2-7B foundation model, yielding high-dimensional representations that capture nucleotide-level biological information. Since the Evo2 embedding space differs from that of the downstream Qwen3 language model, the two modalities are aligned through a learned linear projection,

\begin{equation}
u_{\mathrm{DNA}} = W h_{\mathrm{Evo2}},
\end{equation}

where $h_{\mathrm{Evo2}}$ is the frozen genomic representation, $W$ is a trainable projection matrix, and $u_{\mathrm{DNA}}$ is the projected embedding in the language-model space. This projected embedding is inserted at a dedicated placeholder token (\texttt{<|dna\_pad|>}) in the prompt, allowing Qwen3 to condition autoregressive, open-ended disease prediction on genomic evidence. The model is trained in two stages: supervised fine-tuning (SFT) to align genomic and language representations, followed by Group Relative Policy Optimization (GRPO)~\cite{grpo} to refine the generated reasoning and final prediction through reinforcement learning.

\textit{Effective as it is, BioReason couples genomic and language reasoning through a fundamentally static interface, which exposes three limitations that motivate {\normalfont\bfseries\itshape GenoMorph}. The linear projection is question-agnostic: it produces the same genomic embedding regardless of what is being asked, so sequence evidence cannot be aligned to the semantics of an individual query. Genomic information is injected once and never revisited, preventing any iterative interaction between sequence evidence and intermediate reasoning. And reinforcement optimization acts only on the output, leaving the latent reasoning trajectory unregulated and offering no pressure toward pathway-consistent biology. {\normalfont\bfseries\itshape GenoMorph} addresses these seams in turn, as described next.}

\subsection{{\normalfont\bfseries\itshape GenoMorph} Architecture}
\label{sec:architecture}

BioReason integrates genomic evidence through a single static projection that is computed once and never revisited, and optimizes only the language-model output. This static interface leaves three critical gaps: genomic representations cannot adapt to the question being asked, sequence evidence cannot be re-consulted as reasoning unfolds, and the reasoning trajectory itself is not explicitly optimized for pathway-consistent biology. {\normalfont\bfseries\itshape GenoMorph} addresses these limitations through a four-stage training pipeline that progressively establishes question-aware genomic representations, efficient latent reasoning, controlled evidence reinjection, geometry-aware biological supervision, and self-adaptive reasoning behavior.
 
\begin{figure*}[t]
\centering
\includegraphics[width=\textwidth]{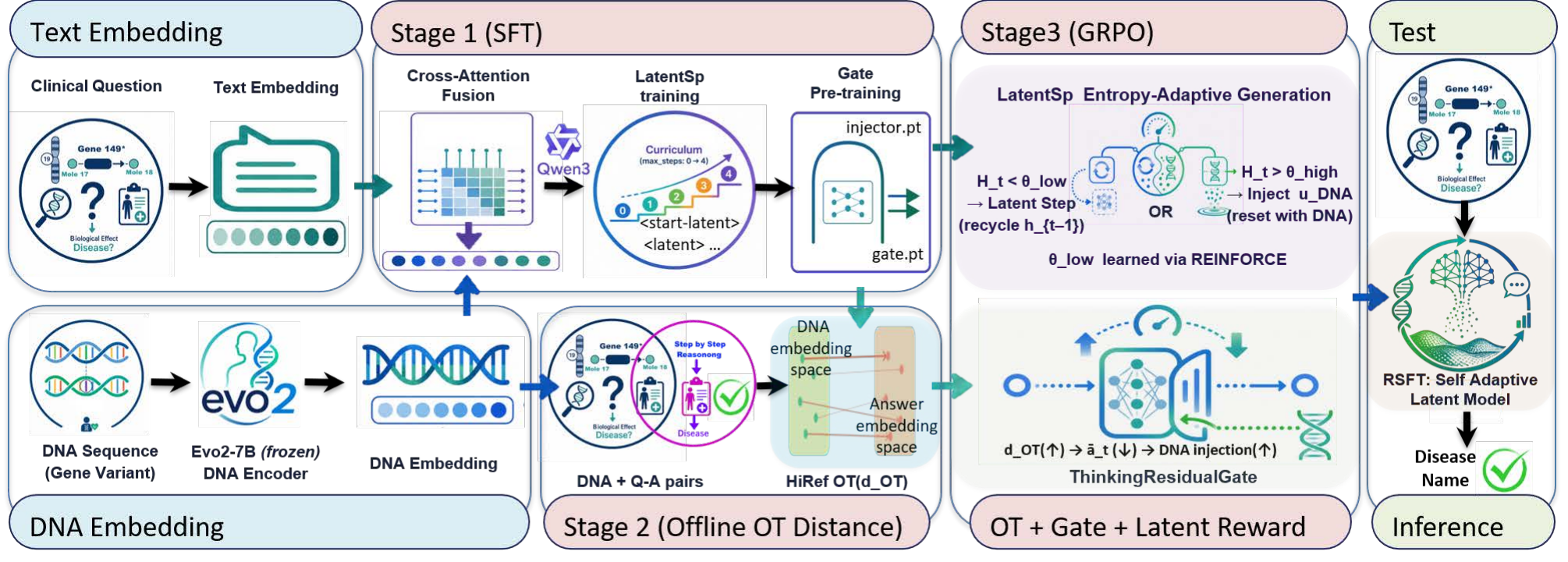}
\caption{\textbf{Training and inference pipeline of {\normalfont\bfseries\itshape GenoMorph}.} The framework begins by encoding genomic DNA sequences using the frozen Evo2-7B foundation model to obtain DNA embeddings, which are combined with clinical question embeddings through question-conditioned CrossAttentionFusion (Stage~1) to generate question-aware genomic representations. Then it performs curriculum-based LatentSp training to initialize latent reasoning, followed by ThinkingResidualGate pretraining for controlled genomic evidence reinjection. Stage~2 computes offline Hierarchical Optimal Transport (HiRef-OT) target manifolds and correspondences (\(d_{\mathrm{OT}}\)), providing geometry-aware rewards for reinforcement optimization. Stage~3 first applies GRPO to optimize latent reasoning trajectories, genomic evidence reinjection, and language generation using the HiRef-based reward. The resulting reasoning traces are subsequently filtered and used for Rejection Sampling Fine-Tuning (RSFT), enabling the model to internalize the adaptive reasoning policy without requiring explicit per-step entropy estimation or a learned entropy threshold during inference. The final self-adaptive latent model dynamically allocates computation by continuing latent reasoning for difficult samples while reinjecting genomic evidence only when necessary, producing open-ended disease predictions grounded in biologically coherent pathway reasoning rather than direct gene--disease associations.}
\label{fig:genomorph_pipeline}
\end{figure*}

\textbf{Stage 1: Supervised Fine-Tuning (SFT)} establishes the core
reasoning mechanism. It combines question-conditioned
\emph{CrossAttentionFusion} for genomic-language alignment,
curriculum-based \emph{LatentSp} for latent reasoning initialization, and \emph{Thinking-ResidualGate} pretraining for controlled genomic evidence reinjection. \textbf{Stage 2: Offline Hierarchical Optimal Transport (HiRef-OT)} freezes the Stage-1 model and estimates the geometric discrepancy between genomic and
disease-answer representations, producing a fixed target manifold and correspondence against which a live optimal transport (OT) distance is measured during reinforcement optimization. \textbf{Stage 3: Group Relative Policy Optimization (GRPO)} jointly optimizes prediction accuracy, reasoning quality, biological consistency, and adaptive latent computation through a multi-objective reward informed by the HiRef
signal. \textbf{Stage 4: Rejection Sampling Fine-Tuning (RSFT)} consolidates the adaptive reasoning behavior learned during GRPO into the model parameters, allowing the final model to execute its learned policy without explicit online entropy estimation during inference. Figure~\ref{fig:genomorph_pipeline} shows the complete pipeline; the following sections describe the four stages in order.
The following subsections describe each component in detail.
 
\subsubsection{Stage 1: Supervised Fine-Tuning}
\label{sec:sft}
Stage 1 establishes the foundational representation and reasoning mechanisms before reinforcement optimization. The stage proceeds through three components: CrossAttentionFusion first converts static genomic representations into question-aware embeddings; LatentSp then initializes the model to compress confident reasoning into latent representations; and ThinkingResidualGate is subsequently pretrained to selectively reinject genomic evidence as reasoning evolves.

\paragraph{CrossAttentionFusion: Question-conditioned Genomic Alignment.}
 
BioReason projects the frozen Evo2 representation through a single linear map, producing the same genomic embedding regardless of the biological question. Consequently, sequence evidence relevant to one reasoning task cannot be selectively emphasized over evidence relevant to another. {\normalfont\bfseries\itshape GenoMorph} instead makes genomic representation a function of the question.
Let
\[
H_{\mathrm{DNA}} \in \mathbb{R}^{B \times S \times d_{\mathrm{DNA}}}
\]
denote the contextual token representations from the frozen Evo2-7B
encoder, where $B$ is the batch size, $S$ the DNA sequence length, and
$d_{\mathrm{DNA}}=4096$, and let
\[
H_Q \in \mathbb{R}^{B \times T \times d_{\mathrm{LLM}}}
\]
denote the embeddings of the complete biological prompt from the Qwen3
embedding layer, where $T$ is the prompt length and
$d_{\mathrm{LLM}}=2048$.
 
\textbf{Cross-attention Formulation:} Genomic representations form the queries while textual embeddings provide
the keys and values,
\[
Q = H_{\mathrm{DNA}}W_Q,\qquad
K = H_QW_K,\qquad
V = H_QW_V,
\]
where $W_Q$, $W_K$, and $W_V$ are learnable projections. The attention
weights are
\[
A=\mathrm{softmax}\left(\frac{QK^\top}{\sqrt{d_h}}\right),
\]
where $d_h$ is the per-head dimension. We use eight attention heads and
mask padding positions before normalization. The question-conditioned
genomic representation is
\[
H'_{\mathrm{DNA}}=W_O(AV),
\]
so that each genomic token selectively attends to the semantically relevant
parts of the query.
 
\textbf{Multimodal integration:} $H'_{\mathrm{DNA}}$ is projected into the Qwen3 hidden dimension and
inserted at the \texttt{<|dna\_pad|>} placeholder positions of the input stream. This preserves token-level genomic granularity while allowing genomic representations to adapt to the reasoning objective. The resulting question-aware embeddings are subsequently used to initialize latent reasoning through LatentSp.
 
\textbf{LatentSp: Curriculum-based Latent Reasoning:} Question-aware representations still leave the cost of explicit autoregressive reasoning: every intermediate biological transition must be generated token by token, even when the model is highly confident about the next reasoning step. LatentSp addresses this inefficiency by progressively moving confident reasoning segments into latent space during supervised training.
 
\textbf{Curriculum Latent Supervision:} Each training sample contains an explicit reasoning trajectory in
\texttt{<think>} tags, divided into reasoning steps. For each token, the
predictive entropy is
\[
H_t=-\sum_{v=1}^{|\mathcal{V}|}p_t(v)\log p_t(v),
\]
and the average entropy of reasoning step $k$ is
\[
H^{(k)}_{\mathrm{step}}
=
\frac{1}{|S_k|}
\sum_{t\in S_k}H_t.
\]
 
Low-entropy steps identify reasoning transitions for which the model is
already confident. These steps are progressively replaced with latent
placeholders,
\[
\texttt{<start-latent>}
\underbrace{\texttt{<latent>}\cdots\texttt{<latent>}}_{m}
\texttt{<end-latent>},
\]
while preserving their original length. The curriculum begins by replacing only the most confident steps and progressively increases the proportion of latent segments, allowing the model to gradually migrate confident reasoning into its hidden state.
 
\textbf{Latent Reasoning Objective:} With $Y$ denoting the original reasoning sequence and $\tilde{Y}$ its
latent-replaced counterpart, the model is trained autoregressively over
the visible tokens,
\[
\mathcal{L}_{\mathrm{LatentSp}}
=
-\sum_{t\in\Omega}
\log P(y_t\mid x,\tilde{y}_{<t}),
\]
where $\Omega$ denotes the set of non-latent positions, each weighted uniformly. This objective encourages the model to preserve reasoning consistency despite omitted intermediate tokens. The resulting latent reasoning policy provides the initial representation for ThinkingResidualGate pretraining.

\textbf{ThinkingResidualGate: Genomic Evidence Reinjection:} As reasoning proceeds through multiple autoregressive or latent transitions, the influence of the original genomic representation can gradually weaken. ThinkingResidualGate is therefore pretrained to selectively reinject genomic evidence into the evolving reasoning state without overwriting the linguistic context.
 
\textbf{Gate Formulation:} Let $E\in\mathbb{R}^{B\times T\times d}$ denote the token embeddings
entering the language model and
$u_{\mathrm{DNA}}\in\mathbb{R}^{B\times d}$ the sequence-level genomic
representation obtained by mean-pooling the CrossAttentionFusion output.
After broadcasting this representation as
$R=\mathrm{Broadcast}(u_{\mathrm{DNA}})$, the gate learns
\[
r=\sigma(W_rE),\qquad
i=\sigma(W_iE),
\]
where $r$ controls residual preservation and $i$ controls genomic evidence
injection. The residual coefficient is
\[
a=r_{\min}+(r_{\max}-r_{\min})\sigma(\lambda+r),
\]
with $r_{\min}=0.5$ and $r_{\max}=0.99$. The gated representation becomes
\[
\tilde{E}
=
aE+
\sqrt{1-a^2+\epsilon}(i\odot R),
\]
thereby preserving linguistic information while selectively restoring genomic evidence.
 
\textbf{Genomic Evidence Reinjection:} During Stage-1 pretraining, the gate learns the basic mechanism for balancing the evolving reasoning state against genomic evidence. Its sample-specific biological grounding is subsequently refined using the HiRef-OT signal computed in Stage 2, allowing reinforcement optimization to preferentially strengthen genomic reinjection for representations that are geometrically less aligned with their corresponding disease-answer representations.
 
\subsubsection{Stage 2: Offline Hierarchical Optimal Transport (HiRef-OT)}
\label{sec:hiref}
 
After Stage 1, the model is frozen and used to construct a geometry-aware supervision signal. Rather than relying only on pairwise similarity measures, {\normalfont\bfseries\itshape GenoMorph} applies hierarchical optimal transport methods such as HiRef~\cite{hiref} to estimate the global discrepancy between the learned genomic representation manifold and the corresponding disease-answer representation manifold.
 
\textbf{Representation Extraction:} For each training sample, the CrossAttentionFusion output is mean-pooled as
\[
u_i^{\mathrm{DNA}}
=
\frac{1}{S}
\sum_{s=1}^{S}H'_{\mathrm{DNA},s},
\]
while the answer representation is mean-pooled over target answer tokens,
\[
u_i^{\mathrm{Ans}}
=
\frac{1}{|A_i|}
\sum_{t\in A_i}h_t^{\mathrm{Ans}}.
\]
 
After $\ell_2$ normalization, the representations form the genomic and
answer manifolds
\[
X=\{\hat{u}_i^{\mathrm{DNA}}\}_{i=1}^{N},
\qquad
Y=\{\hat{u}_i^{\mathrm{Ans}}\}_{i=1}^{N}.
\]
 
\textbf{Hierarchical Monge Mapping:} Using the transport cost
\[
C_{ij}
=
\left\|
\hat{u}_i^{\mathrm{DNA}}
-
\hat{u}_j^{\mathrm{Ans}}
\right\|_2^2,
\]
HiRef estimates a correspondence
\[
T^*
=
\arg\min_T
\sum_{i=1}^{N}C(i,T(i)).
\]
The hierarchical procedure recursively decomposes the transport problem into smaller subproblems while preserving global geometric consistency, providing an efficient approximation to the full Monge mapping.
 
\textbf{Offline Target Manifold and Live OT Distance:} The Monge correspondence $T^*$ and the resulting answer-manifold targets are computed once from the frozen Stage-1 model and stored offline. What is fixed, in other words, is the \emph{target geometry} — the answer embeddings and the per-sample target assignment — not the distance itself. During subsequent reinforcement optimization the genomic backbone and gate continue to train, so the genomic representation $\hat{u}_i^{\mathrm{DNA}}$ moves; the per-sample discrepancy
\[
d_{\mathrm{OT}}^{(i)}
=
\left\|
\hat{u}_i^{\mathrm{DNA}}
-
\hat{u}_{T^*(i)}^{\mathrm{Ans}}
\right\|_2^2
\]
is therefore recomputed live at every optimization step against the fixed target, rather than cached as a static value.
 
The fixed target manifold and Monge assignment provide a geometry-aware signal for both the multi-objective GRPO reward and the adaptive genomic evidence reinjection mechanism.
 
\subsubsection{Stage 3: Group Relative Policy Optimization
(GRPO)}
\label{sec:grpo}
 
The first two stages establish question-aware genomic representations, latent reasoning, controlled evidence reinjection, and a fixed target manifold for measuring biological-geometric consistency. Stage 3 jointly optimizes these capabilities using Group Relative Policy Optimization (GRPO), encouraging the model to generate trajectories that are accurate, biologically coherent, geometrically consistent, and computationally efficient.
 
\textbf{Multi-objective Reward:} For each input $x$, the policy samples a group of $G$ trajectories
$Y=\{y_1,\ldots,y_G\}$. Each trajectory receives the eight-term weighted reward
\begin{align*}
R(y)
={}& 0.5\,R_{\mathrm{format}}
+ 2.0\,R_{\mathrm{correct}}
+ 0.5\,R_{\mathrm{compqual}} \\
& + 0.3\,R_{\mathrm{reasqual}}
+ 0.5\,R_{\mathrm{latentfmt}}
+ 0.5\,R_{\mathrm{OT}} \\
& + 0.75\,R_{\mathrm{latent}}
+ 0.5\,R_{\mathrm{concise}},
\end{align*}
with weights as used in training. $R_{\mathrm{format}}$ is a single unified structural-format score (valid \texttt{<think>}/answer tagging), not a sum of separate sub-scores. $R_{\mathrm{correct}}$ rewards prediction correctness and receives by far the largest weight. $R_{\mathrm{compqual}}$ and $R_{\mathrm{reasqual}}$ separately score completion quality and reasoning-diversity (repetition avoidance), and are kept as distinct terms with distinct weights rather than merged, since they are weighted differently (0.5 vs.\ 0.3) in training. $R_{\mathrm{latentfmt}}$ penalizes malformed latent blocks (a \texttt{<start-latent>} span left unclosed by \texttt{<end-latent>}). $R_{\mathrm{OT}}$ rewards consistency with the HiRef-derived geometric signal. $R_{\mathrm{latent}}$ rewards efficient latent computation. $R_{\mathrm{concise}}$ is a separate brevity signal over step count and total length, distinct from $R_{\mathrm{reasqual}}$.
 
\textbf{Group-relative optimization:} Each sampled trajectory is evaluated relative to the mean reward of its
group,
\[
A_i
=
R(y_i)
-
\frac{1}{G}\sum_{j=1}^{G}R(y_j),
\]
and the policy is optimized according to a clipped, KL-regularized surrogate objective in the style of PPO,
\[
\mathcal{L}_{\mathrm{GRPO}}
=
\mathbb{E}
\left[
\min\!\left(\rho_i A_i,\ \mathrm{clip}(\rho_i,1-\epsilon,1+\epsilon)A_i\right)
-\beta\,D_{\mathrm{KL}}\!\left(\pi_\theta\,\|\,\pi_{\mathrm{ref}}\right)
\right],
\]
where the importance ratio is
\[
\rho_i=\frac{\pi_\theta(y_i\mid x)}{\pi_{\theta_{\mathrm{old}}}(y_i\mid x)},
\]
with clipping range $\epsilon=0.1$ and KL coefficient $\beta=0.05$, matching the configuration used by TRL's \texttt{GRPOTrainer} in our training runs. This relative objective favors trajectories that better integrate genomic evidence and pathway-consistent reasoning than alternative solutions generated for the same input.
 
\textbf{Adaptive Latent Computation:} GRPO also governs when further latent computation is beneficial, controlled by a threshold parameter $\theta_{\mathrm{low}}$ on step-level entropy. A naive baseline treats $\theta_{\mathrm{low}}$ as a fixed, hand-set target that is linearly ramped over training by a warmup schedule, so that the computation budget is manually fixed rather than learned. {\normalfont\bfseries\itshape GenoMorph} instead optimizes $\theta_{\mathrm{low}}$ online via a derivative-free sliding-window search: at each update, $\theta_{\mathrm{low}}$ is perturbed in both directions ($\theta_{\mathrm{low}}\pm\delta$), the resulting trajectories are scored, and a finite-difference estimate of the reward gradient against a windowed reward baseline is used to update $\theta_{\mathrm{low}}$, subject to a fixed step size and clipping range. We deliberately avoid a REINFORCE-style policy-gradient update for this parameter, $\theta \leftarrow \theta+\eta\,\nabla_\theta\log\pi(\text{decision})\cdot A$, because it proved uninformative in this setting: GRPO's per-group advantage $A_i$ is zero-mean by construction, and the latent-continuation decision is shared across every completion in a group, so no per-example credit signal survives to update $\theta_{\mathrm{low}}$. The sliding-window search sidesteps this by evaluating $\theta_{\mathrm{low}}$'s effect directly against a windowed reward baseline rather than relying on per-trajectory credit assignment, and it is this mechanism, not a fixed schedule or REINFORCE, that produces the adaptive threshold used in the reported model.
 
The resulting policy learns to allocate additional computation to difficult genomic reasoning problems while avoiding unnecessary latent transitions for simpler cases. However, these adaptive decisions still depend on explicit uncertainty estimation during inference, motivating the final RSFT stage.

\subsubsection{Stage 4: Rejection Sampling Fine-Tuning (RSFT)}
\label{sec:rsft}

Although Stage 3 learns an adaptive reasoning policy through GRPO, not all sampled trajectories are equally useful for subsequent training. Some may produce incorrect predictions, malformed output, or unnecessarily long latent computation despite being generated by the optimized policy. Stage 4 therefore applies Rejection Sampling Fine-Tuning (RSFT) to construct a high-quality training corpus by filtering candidate trajectories with a two-step selection rule and fine-tuning on the surviving trajectories, reinforcing the successful adaptive reasoning behavior learned during GRPO.
 
\textbf{Candidate Trajectory Generation:} After GRPO converges, the optimized policy generates multiple candidate
trajectories for each training instance. For an input $x$, let
\[
\mathcal{Y}_x=\{y_1,y_2,\ldots,y_G\}
\]
denote a group of $G$ sampled trajectories. Each trajectory contains the
reasoning process, latent transitions, genomic evidence reinjection events,
and final prediction.
 
\textbf{Selection: a two-step filter, not a scalar reward threshold.} Trajectories are not ranked by a weighted sum of reward terms. Instead, selection proceeds in two steps:
\begin{enumerate}
\item \textbf{Hard gate.} A trajectory is admissible only if it is correct \emph{and} well-formed — that is, it parses into a valid reasoning structure with a non-empty final answer. Trajectories failing either condition are discarded outright; no OT, format, or reasoning-quality score is consulted at this stage.
\item \textbf{Preference among survivors.} Among admissible trajectories for a given input, a trajectory that uses latent reasoning is preferred over one that does not; ties are broken by shortest completion (fewest tokens), or, when timing-based preference is enabled, by lowest generation time.
\end{enumerate}
Formally, writing $\mathrm{ok}(y)$ for the correctness-and-well-formedness gate, $\mathrm{lat}(y)\in\{0,1\}$ for whether $y$ uses latent reasoning, and $\mathrm{len}(y)$ for completion length, the accepted trajectory for input $x$ is
\[
y^*(x)
=
\operatorname*{arg\,max}_{y\in\mathcal{Y}_x\,:\,\mathrm{ok}(y)=1}
\big(\mathrm{lat}(y),\ -\mathrm{len}(y)\big),
\]
i.e., \ a lexicographic selection — filter to admissible trajectories, then prefer latent-using ones, then prefer shorter ones — rather than a threshold or arg\,max over a scalar multi-objective reward.
\[
\mathcal{D}_{\mathrm{RSFT}}
=
\{(x,y^*(x))\mid x\in\mathcal{X}\}.
\]
 
Because the targets already encode the adaptive decisions discovered during reinforcement optimization, the model progressively internalizes the reasoning policy into its parameters rather than explicitly evaluating a threshold at every generation step.
 
\textbf{Fine-tuning on accepted trajectories:}
 
The accepted trajectories are subsequently used as supervised targets for
fine-tuning. For an accepted trajectory
\[
Y=(y_1,\ldots,y_T),
\]
the RSFT objective is
\[
\mathcal{L}_{\mathrm{RSFT}}
=
-\sum_{t=1}^{T}
\log P_\theta(y_t\mid x,y_{<t}).
\]
 
Because the training corpus contains only trajectories that exhibit the
desired reasoning behavior, RSFT reinforces the adaptive latent decisions
and genomic evidence reinjection patterns discovered during GRPO while
discarding unsuccessful alternatives. Consequently, the model learns to
prefer high-quality reasoning trajectories without being repeatedly exposed
to low-reward or biologically inconsistent generations.
 
\textbf{Self-adaptive Inference.} At inference, the final model operates without explicit per-step entropy estimation or manually maintained decision thresholds. Difficult samples can naturally trigger longer latent reasoning trajectories, whereas simpler samples terminate earlier. When the evolving reasoning state requires additional biological grounding, the pretrained residual mechanism selectively revisits genomic evidence. The resulting self-adaptive policy therefore combines efficient computation with pathway-grounded genomic reasoning, reducing reliance on direct gene--disease associations.

\subsection{Training configuration, evaluation metrics, and baselines}
 
\paragraph{Implementation details.}
%%%%%%%%%%%%%%%%%%%%%%%%%%%%%%%%%%%%%%%%%%%%%%

All experiments were conducted using the {\normalfont\bfseries\itshape GenoMorph} framework described in
Section~\ref{sec:architecture}. Evo2-7B was employed as the frozen genomic foundation model to encode DNA sequences, while Qwen3-1.7B \cite{qwen3} served as the language backbone. CrossAttentionFusion uses eight attention heads to dynamically align genomic sequence representations with the semantic context of the biological query before multimodal
integration. Trainable components are combined with LoRA adapters \cite{lora} throughout supervised training, GRPO, and RSFT (see Tables~\ref{tab:sft-config}, \ref{tab:grpo-config}, and
\ref{tab:rft-config} for the specific rank and $\alpha$ settings used in each stage).
 
Training follows the four sequential stages introduced in
Section~\ref{sec:architecture}: (1) supervised optimization of
CrossAttentionFusion in place of BioReason's static linear projection;
(2) LatentSp curriculum learning and ThinkingResidualGate pretraining;
(3) offline computation of hierarchical optimal transport (HiRef-OT) distances between genomic and answer manifolds; and (4) Group RelativePolicy Optimization (GRPO), followed by Rejection Sampling Fine-Tuning (RSFT), which distills the GRPO-learned adaptive latent policy into the final inference model without requiring online entropy computation. The principal hyperparameters used during supervised learning, GRPO optimization, and Rejection Sampling Fine-Tuning are summarized in
Tables~\ref{tab:sft-config}, \ref{tab:grpo-config}, and
\ref{tab:rft-config}, respectively. Unless otherwise stated, all results reported in Section~\ref{sec:results} correspond to the final RSFT model.
%%%%%%%%%%%%%%%%%%%%%%%%%%%%%%%%%%%%%%%
 
\begin{table}[htbp]
  \centering
  \caption{Training configuration for supervised learning (Stage 1).}
  \label{tab:sft-config}
  \begin{tabular}{lp{5cm}}
    \toprule
    \textbf{Component} & \textbf{Setting} \\
    \midrule
    Backbone LLM                & Qwen3-1.7B-Instruct \\
    Genomic encoder             & Evo2-7B (frozen) \\
    DNA embedding layer         & \texttt{blocks.28.mlp.l3} \\
    Fusion module                & CrossAttentionFusion \\
    Trainable parameters        & CrossAttentionFusion + LoRA adapters \\
    LoRA rank / $\alpha$ / dropout & 32 / 64 / 0.05 \\
    Optimizer                   & AdamW \\
    Learning rate                & $5 \times 10^{-5}$ \\
    Weight decay                 & 0.01 \\
    Learning rate schedule       & Cosine decay (10\% warmup) \\
    Batch size                   & 1 (effective batch size = 8) \\
    Training epochs              & 5 (best checkpoint selected on validation set) \\
    Maximum text length          & 6000 tokens \\
    Maximum DNA length            & 2048 bp \\
    DNA preprocessing            & $\pm 1024$ bp around the variant \\
    Precision                    & BF16 mixed precision \\
    Random seed                  & 23 \\
    Hardware                     & NVIDIA A100 80\,GB GPU \\
    Dataset                      & KEGG (original and anonymized) \\
    \bottomrule
  \end{tabular}
\end{table}
 
\begin{table}[htbp]
  \centering
  \caption{Training configuration for Group Relative Policy Optimization (GRPO) (Stage 3).}
  \label{tab:grpo-config}
  \begin{tabular}{lp{5cm}}
    \toprule
    \textbf{Parameter} & \textbf{Value} \\
    \midrule
    Initialization               & Best supervised checkpoint \\
    Optimizer                    & AdamW \\
    Learning rate                & $2 \times 10^{-6}$ \\
    Batch size                   & 1 (effective = 4) \\
    LoRA ($r$, $\alpha$)          & 16, 32 \\
    Generations                  & 8 \\
    Maximum response length       & 800 tokens \\
    Sampling                     & Temperature = 0.7, Top-$p$ = 0.95, Top-$k$ = 50 \\
    Initial entropy thresholds    & $\theta_{\text{low}} = 1.0$, $\theta_{\text{high}} = 3.0$ \\
    Maximum latent steps          & 1 \\
    ThinkingResidualGate          & Enabled \\
    HiRef-OT alignment            & Enabled \\
    Reward components             & Correctness, Format, Completion Quality, Reasoning Quality, Latent Format, HiRef-OT, Latent Usage, Conciseness \\
    OT reward weight              & 0.05 \\
    Gradient clipping             & 0.1 \\
    Precision                     & BF16 mixed precision \\
    Hardware                      & NVIDIA A100 (80\,GB) \\
    \bottomrule
  \end{tabular}
\end{table}
 
\begin{table}[htbp]
  \centering
  \caption{Rejection Sampling fine-tuning (RSFT) configuration.}
  \label{tab:rft-config}
  \begin{tabular}{lp{5cm}}
    \toprule
    \textbf{Parameter} & \textbf{Value} \\
    \midrule
    Initialization                 & Best GRPO checkpoint (LoRA merged) \\
    Optimizer                      & AdamW \\
    Learning rate                  & $1 \times 10^{-5}$ \\
    Weight decay                   & 0.01 \\
    Learning rate schedule          & Cosine decay (5\% warmup) \\
    Batch size                     & 1 (effective = 8) \\
    Training epochs                & 4 (best checkpoint selected on validation set) \\
    Maximum text length             & 6000 tokens \\
    Maximum DNA length               & 2048 bp \\
    DNA preprocessing               & $\pm 1024$ bp around the variant \\
    CrossAttentionFusion            & Enabled \\
    ThinkingResidualGate            & Enabled \\
    HiRef-OT alignment              & Enabled \\
    Training traces                 & GRPO reasoning traces \\
    Adaptive latent reasoning        & Enabled \\
    Online entropy computation       & Disabled \\
    Gradient clipping               & 0.1 \\
    Precision                       & FP16 mixed precision \\
    Hardware                        & NVIDIA A100 (80\,GB) \\
    \bottomrule
  \end{tabular}
\end{table}
 
\paragraph{Evaluation metrics.}
We evaluate {\normalfont\bfseries\itshape GenoMorph} on both the original and anonymized KEGG benchmarks
using prediction accuracy, precision, recall, macro-F1 score,
weighted-F1 score, and average inference time per sample. Accuracy
measures overall disease prediction performance, while macro-F1 evaluates
balanced performance across disease categories and weighted-F1 accounts
for class imbalance. Precision and recall provide complementary measures
of prediction reliability, and inference time quantifies the
computational efficiency of the proposed adaptive reasoning framework.
 
Following BioReason, generated disease names are normalized before
comparison with the ground-truth labels. Performance on the original
KEGG benchmark measures conventional genomic disease prediction, whereas
evaluation on the anonymized benchmark assesses mechanistic reasoning by
removing explicit biological identifiers while preserving genomic
sequences and pathway topology.
 
Prediction accuracy is computed as
\begin{equation}
  \text{Accuracy} = \frac{1}{N} \sum_{i=1}^{N} \mathbb{1}(\hat{y}_i = y_i) ,
  \label{eq:accuracy}
\end{equation}
where $N$ denotes the number of test samples, $y_i$ is the ground-truth
disease label, and $\hat{y}_i$ is the predicted label.
 
Precision and recall for each disease class are computed as
\begin{equation}
  P_c = \frac{TP_c}{TP_c + FP_c} ,
  \label{eq:precision}
\end{equation}
\begin{equation}
  R_c = \frac{TP_c}{TP_c + FN_c} ,
  \label{eq:recall}
\end{equation}
where $TP_c$, $FP_c$, and $FN_c$ denote the true positives, false
positives, and false negatives for class $c$, respectively.
 
The macro-F1 score is defined as
\begin{equation}
  \text{Macro-F1} = \frac{1}{C} \sum_{c=1}^{C} \frac{2 P_c R_c}{P_c + R_c} ,
  \label{eq:macro-f1}
\end{equation}
where $C$ is the total number of disease classes.
 
The weighted-F1 score is computed as
\begin{equation}
  \text{Weighted-F1} = \sum_{c=1}^{C} \frac{n_c}{N} \, F1_c ,
  \label{eq:weighted-f1}
\end{equation}
where $n_c$ is the number of samples belonging to class $c$.
 
Inference efficiency is measured as the average wall-clock time required
to generate one complete reasoning trajectory and the corresponding
disease prediction.
 
\paragraph{Baselines}
We compare {\normalfont\bfseries\itshape GenoMorph} against representative models spanning three categories: general-purpose large language models, biomedical language
models, and multimodal genomic reasoning frameworks. All methods are evaluated using identical training, validation, and test splits on both the original and anonymized KEGG benchmarks to ensure a fair comparison.
 
The first category consists of general-purpose LLMs, including GPT-4o, Gemini, and Qwen3-1.7B, which evaluate the capability of state-of-the-art
language models to perform genomic disease reasoning without explicit genomic representation learning. The second category includes biomedical
language models, namely BioMedGPT, BioMistral, and Meditron, which have been pretrained or instruction-tuned using biomedical literature and
clinical corpora. The third category consists of multimodal genomic reasoning frameworks represented by BioReason, which integrates Evo2-derived genomic representations with Qwen3 through a static linear
projection.
 
For BioReason, we report both the supervised fine-tuning (SFT) model and the GRPO-optimized model. For {\normalfont\bfseries\itshape GenoMorph}, we report the Stage-3 GRPO model
as well as the final rejection sampling fine-tuned (RSFT) model, where adaptive latent reasoning is distilled into the network to eliminate onlinenentropy computation during inference. The baseline models are summarized in Table~\ref{tab:baselines}.
 
\begin{table}[t]
\centering
\caption{Baseline methods used for comparison.}
\label{tab:baselines}
\small
\begin{tabular}{p{1.2cm}p{2.11cm}p{1.7cm}p{2.85cm}}
\toprule
\textbf{Category} &
\textbf{Model} &
\textbf{Input} &
\textbf{Setting} \\
\midrule

General &
GPT-4o &
Text &
Zero-shot \\

General &
Gemini &
Text &
Zero-shot \\

General &
Qwen3-1.7B &
Text &
Zero-shot \\

Biomedical &
BioMedGPT &
Text &
Zero-shot \\

Biomedical &
BioMistral &
Text &
Zero-shot \\

Biomedical &
Meditron &
Text &
Zero-shot \\

Genomic &
BioReason (SFT) &
DNA + Text &
SFT + Linear projection \\

Genomic &
BioReason (GRPO) &
DNA + Text &
GRPO + Linear projection \\

\midrule

\textbf{Proposed} &
{\normalfont\bfseries\itshape GenoMorph} &
\textbf{DNA + Text + HiRef-OT} &
\textbf{CrossAttention + Latent + Gate + GRPO} \\

\textbf{Proposed} &
\textbf{\textit{GenoMorph+}} &
\textbf{DNA + Text} &
\textbf{CrossAttention + Latent + Gate + RSFT} \\

\bottomrule
\end{tabular}
\end{table}

\section{Results} \label{sec:results}
This section evaluates {\normalfont\bfseries\itshape GenoMorph} from three complementary perspectives: (i) predictive performance on both the original and anonymized KEGG benchmarks, (ii) the contribution of each architectural component through extensive ablation studies, and (iii) reasoning efficiency and mechanistic robustness. We first present comparisons with existing genomic reasoning frameworks and large language models, and then analyze how each proposed component contributes to pathway-guided genomic reasoning.

\subsection{Overall Performance on Genomic Disease Reasoning}

\begin{figure}[t]
\centering
\includegraphics[width=\columnwidth]{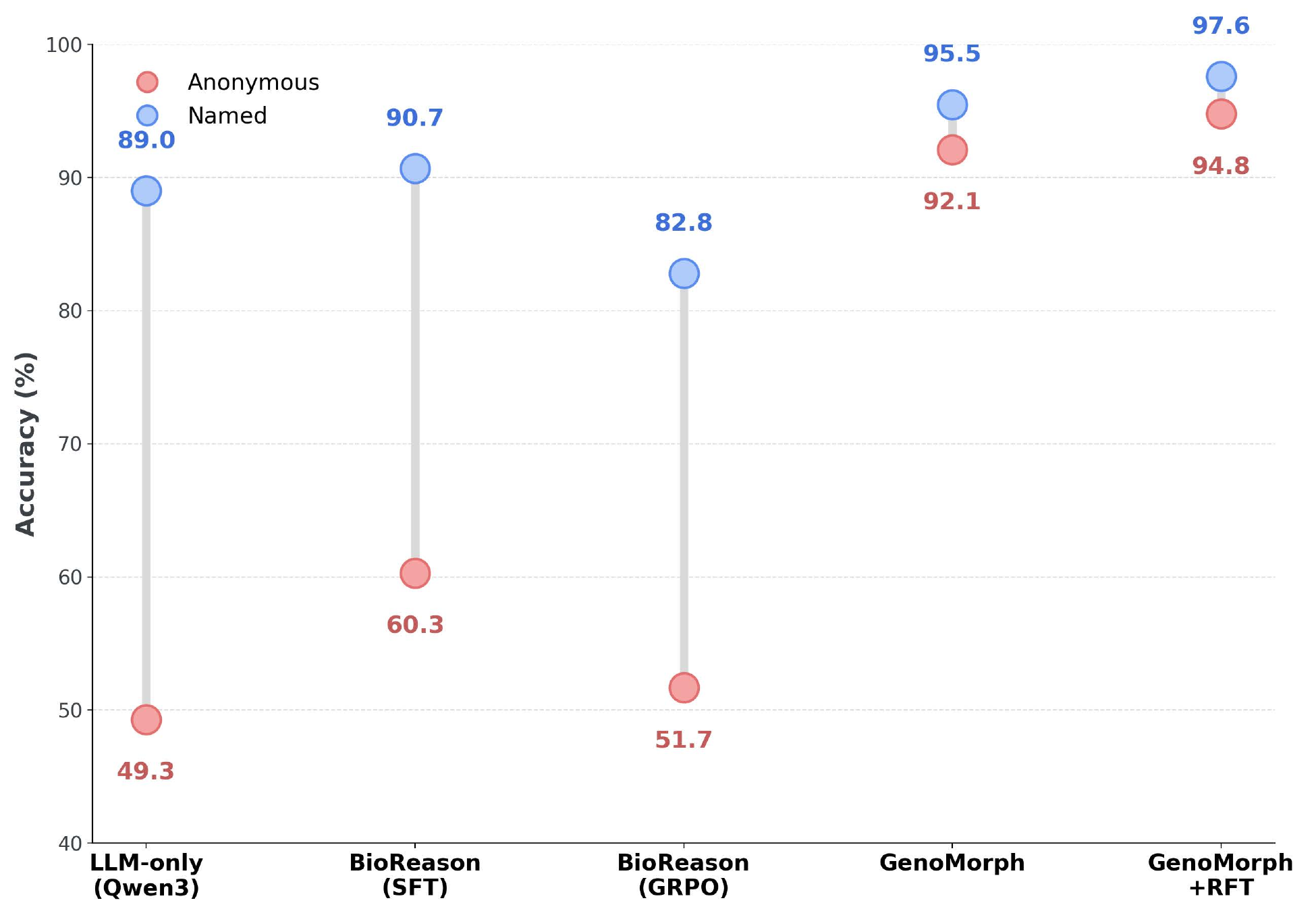}
\caption{Accuracy comparison on the named and anonymized genomic disease reasoning benchmarks. The vertical distance between the paired markers represents the performance degradation caused by removing explicit gene and molecular entity names. {\normalfont\bfseries\itshape GenoMorph} exhibits the smallest performance gap, demonstrating robust sequence-level genomic reasoning under entity anonymization.}
\label{fig:accuracy_comparison}
\end{figure}

\begin{figure}[t]
\centering
\includegraphics[width=\columnwidth]{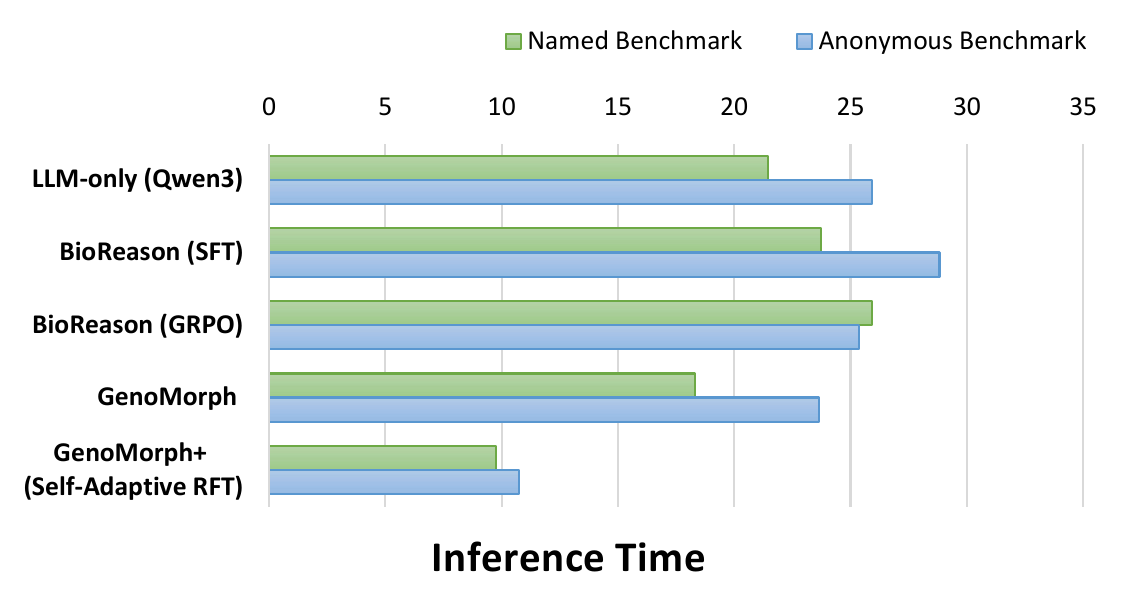}
\caption{Average inference time per sample on the named and anonymized genomic disease reasoning benchmarks. Lower inference time indicates higher computational efficiency. {\normalfont\bfseries\itshape GenoMorph} with self-adaptive RSFT achieves the lowest inference latency while simultaneously providing the highest predictive performance.}
\label{fig:inference_time}
\end{figure}

Table~\ref{tab:main_results} summarizes the overall performance of {\normalfont\bfseries\itshape GenoMorph} and competing methods on the genomic disease reasoning benchmark. Results are reported on both the original benchmark containing explicit gene identifiers (Named) and the anonymized benchmark in which all gene and molecular entity names are replaced with globally consistent anonymous identifiers. In addition to predictive performance, we report the average inference time per sample. The corresponding accuracy and inference-time comparisons are illustrated in Figures~\ref{fig:accuracy_comparison} and~\ref{fig:inference_time}, respectively.

The LLM-only Qwen3 baseline achieved an accuracy of 88.97\% on the named benchmark, demonstrating that modern language models can effectively exploit memorized biological knowledge when explicit gene identifiers are available. However, its performance decreased substantially to 49.31\% after anonymization, indicating a strong dependence on lexical associations between gene names and biological functions. As illustrated in Figure~\ref{fig:accuracy_comparison}, this model exhibits the largest performance gap between the two benchmark settings.

Incorporating genomic sequence representations through BioReason improves disease prediction compared with the LLM-only baseline. The supervised fine-tuned (SFT) model achieved 90.69\% accuracy on the named benchmark and 60.34\% on the anonymized benchmark, while the GRPO variant obtained 82.75\% and 51.72\%, respectively. Although genomic information improves robustness, both variants still experience considerable degradation after anonymization, suggesting that static genomic fusion remains insufficient for fully aligning sequence-derived biological evidence with language reasoning.

{\normalfont\bfseries\itshape GenoMorph} consistently achieves the strongest predictive performance across both benchmark settings. The proposed framework attains 95.52\% accuracy on the named benchmark together with macro- and weighted-F1 scores of 0.846 and 0.941, respectively. More importantly, it maintains high performance on the anonymized benchmark, achieving 92.07\% accuracy with a weighted-F1 score of 0.929. As shown in Figure~\ref{fig:accuracy_comparison}, the performance gap between the named and anonymized benchmarks is substantially smaller than that of all competing approaches, indicating that {\normalfont\bfseries\itshape GenoMorph} relies primarily on sequence-derived biological representations rather than memorized gene identities.

The proposed self-adaptive Rejection Sampling Fine-Tuning (RSFT) further improves both predictive performance and computational efficiency. The final model achieves the highest accuracy of 97.59\% on the named benchmark and 94.83\% on the anonymized benchmark while simultaneously reducing the average inference time from 18.31~s to 9.75~s on the named benchmark and from 23.64~s to 10.73~s on the anonymized benchmark. Figure~\ref{fig:inference_time} shows that the RSFT model requires the lowest inference time among all evaluated approaches while also delivering the highest prediction accuracy, demonstrating that self-adaptive latent reasoning improves both effectiveness and efficiency.

General-purpose and biomedical language models evaluated in the zero-shot setting exhibit limited genomic reasoning capability. Although GPT-4o and Gemini produce moderate performance on the named benchmark, both experience severe degradation after anonymization. Domain-specific language models, including BioMedGPT, BioMistral, and Meditron, perform poorly on both benchmark variants. These observations suggest that pretrained biomedical knowledge alone is insufficient for robust genomic disease reasoning without explicit integration of genomic sequence representations.

\begin{table*}[t]
\centering
\caption{Overall performance on the genomic disease reasoning benchmark. Results are reported on both the original benchmark (Named) and the anonymized benchmark (Anonymous). Macro-F1 (M), Weighted-F1 (W), and average inference time per example are reported. Best results are shown in \textbf{bold}.}
\label{tab:main_results}
%\resizebox{\textwidth}{!}{
\begin{tabular}{lcccccccc}
\toprule
\multirow{2}{*}{\textbf{Model}} &
\multicolumn{4}{c}{\textbf{Named Benchmark}} &
\multicolumn{4}{c}{\textbf{Anonymous Benchmark}} \\
\cmidrule(lr){2-5}\cmidrule(lr){6-9}
& Acc. & Macro-F1 & Weighted-F1 & Time (s)
& Acc. & Macro-F1 & Weighted-F1 & Time (s) \\
\midrule

\multicolumn{9}{c}{\textbf{Trained Models}}\\
\midrule

LLM-only (Qwen3)
& 0.8897 & 0.5959 & 0.7461 & 21.45
& 0.4931 & 0.1921 & 0.4256 & 25.93 \\

BioReason (SFT)
& 0.9069 & 0.7700 & 0.7863 & 23.74
& 0.6034 & 0.3798 & 0.6032 & 28.83 \\

BioReason (GRPO)
& 0.8275 & 0.6828 & 0.8550 & 25.91
& 0.5172 & 0.2176 & 0.4812 & 25.37 \\
\rowcolor{blue!20}
\textbf{\textit{GenoMorph} (Ours)}
& 0.9552 & 0.8461 & 0.9412 & 18.31
& 0.9207 & 0.8266 & 0.9285 & 23.64 \\

\textbf{\makecell[l]{\textit{GenoMorph+}\\(Self-Adaptive RSFT)}}
& \textbf{0.9759}
& \textbf{0.9449}
& \textbf{0.9725}
& \textbf{9.75}
& \textbf{0.9483}
& \textbf{0.8358}
& \textbf{0.9465}
& \textbf{10.73} \\

\midrule
\multicolumn{9}{c}{\textbf{Zero-shot Models}}\\
\midrule

GPT-4o
& 0.614 & 0.162 & 0.162 & API
& 0.128 & 0.018 & 0.018 & API \\

Gemini
& 0.703 & 0.254 & 0.254 & API
& 0.086 & 0.035 & 0.035 & API \\

BioMedGPT
& 0.062 & 0.028 & 0.028 & 27.14
& 0.000 & 0.000 & 0.000 & 26.74 \\

BioMistral
& 0.062 & 0.014 & 0.014 & 02.34
& 0.000 & 0.000 & 0.000 & 02.15 \\

Meditron
& 0.003 & 0.003 & 0.003 & 19.89
& 0.000 & 0.000 & 0.000 & 19.42 \\

\bottomrule
\end{tabular}
%}
\end{table*}

%%%%%%%%%%%%%%%%%%%%%%%%%%%%%%%%%%%%%%%%%%%%%%%%%%%%%%%

\subsection{Performance under Gene-name Anonymization}

\begin{figure}[t]
\centering
\includegraphics[width=\columnwidth]{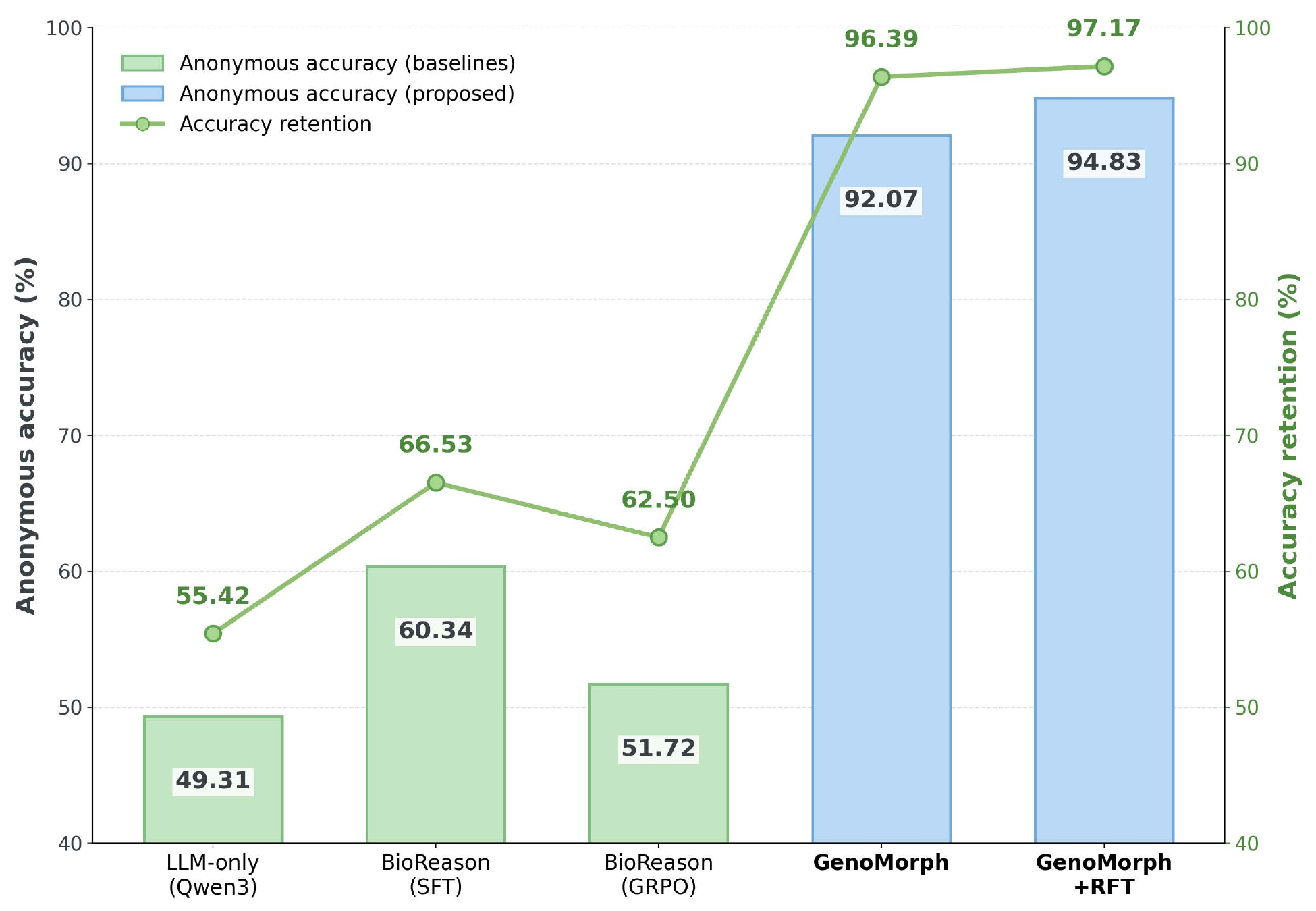}
\caption{Anonymous benchmark accuracy and accuracy retention after gene-name anonymization. Gray bars denote baseline models, whereas blue bars correspond to the proposed {\normalfont\bfseries\itshape GenoMorph} variants. The red line represents accuracy retention relative to the named benchmark. {\normalfont\bfseries\itshape GenoMorph} and {\normalfont\bfseries\itshape GenoMorph}+RSFT achieve both substantially higher anonymous accuracy and significantly greater accuracy retention, demonstrating robust sequence-level genomic reasoning under entity anonymization.}
\label{fig:performance_retention}
\end{figure}

Table~\ref{tab:anonymization_drop} summarizes the robustness of each model after all gene and molecular entity names were replaced with globally consistent anonymous identifiers, while Figure~\ref{fig:performance_retention} visualizes the anonymous benchmark accuracy together with the corresponding accuracy retention. This evaluation isolates a model's ability to reason from genomic sequence information rather than relying on memorized associations between gene names and biological functions.

The LLM-only Qwen3 baseline exhibits the largest degradation, losing 39.66 percentage points in accuracy and retaining only 55.4\% of its original performance. Incorporating DNA sequence representations through BioReason improves robustness, increasing accuracy retention to 66.5\% after supervised fine-tuning, although substantial degradation remains.

In contrast, {\normalfont\bfseries\itshape GenoMorph} shows only a 3.45-point reduction in accuracy, preserving 96.4\% of its original performance. The proposed self-adaptive RSFT further improves robustness, reducing the accuracy loss to just 2.76 percentage points while retaining 97.2\% of the original accuracy. These results demonstrate that {\normalfont\bfseries\itshape GenoMorph} primarily reasons from genomic sequence representations instead of memorized biological terminology, enabling consistent performance even when explicit gene identifiers are unavailable.

\begin{table}[t]
\centering
\caption{Robustness to gene-name anonymization. Absolute decrease in accuracy and accuracy retention after replacing all gene and molecular entity names with globally consistent anonymous identifiers. Lower accuracy drops and higher retention indicate stronger sequence-level genomic reasoning.}
\label{tab:anonymization_drop}
\begin{tabular}{lcc}
\toprule
\textbf{Model} &
\textbf{$\Delta$ Accuracy $\downarrow$} &
\textbf{Accuracy Retention (\%) $\uparrow$} \\
\midrule
LLM-only (Qwen3)         & 39.66 & 55.4 \\
BioReason (SFT)          & 30.35 & 66.5 \\
BioReason (GRPO)         & 31.03 & 62.5 \\
\rowcolor{blue!20}
\textbf{\textit{GenoMorph} (Ours)} & 3.45 & 96.4 \\
\textbf{\makecell[l]{\textit{GenoMorph+}\\(Self-Adaptive RSFT)}} & \textbf{2.76} & \textbf{97.2} \\
\bottomrule
\end{tabular}
\end{table}

\subsection{Ablation Study}

\begin{figure}[t]
\centering
\includegraphics[width=\columnwidth]{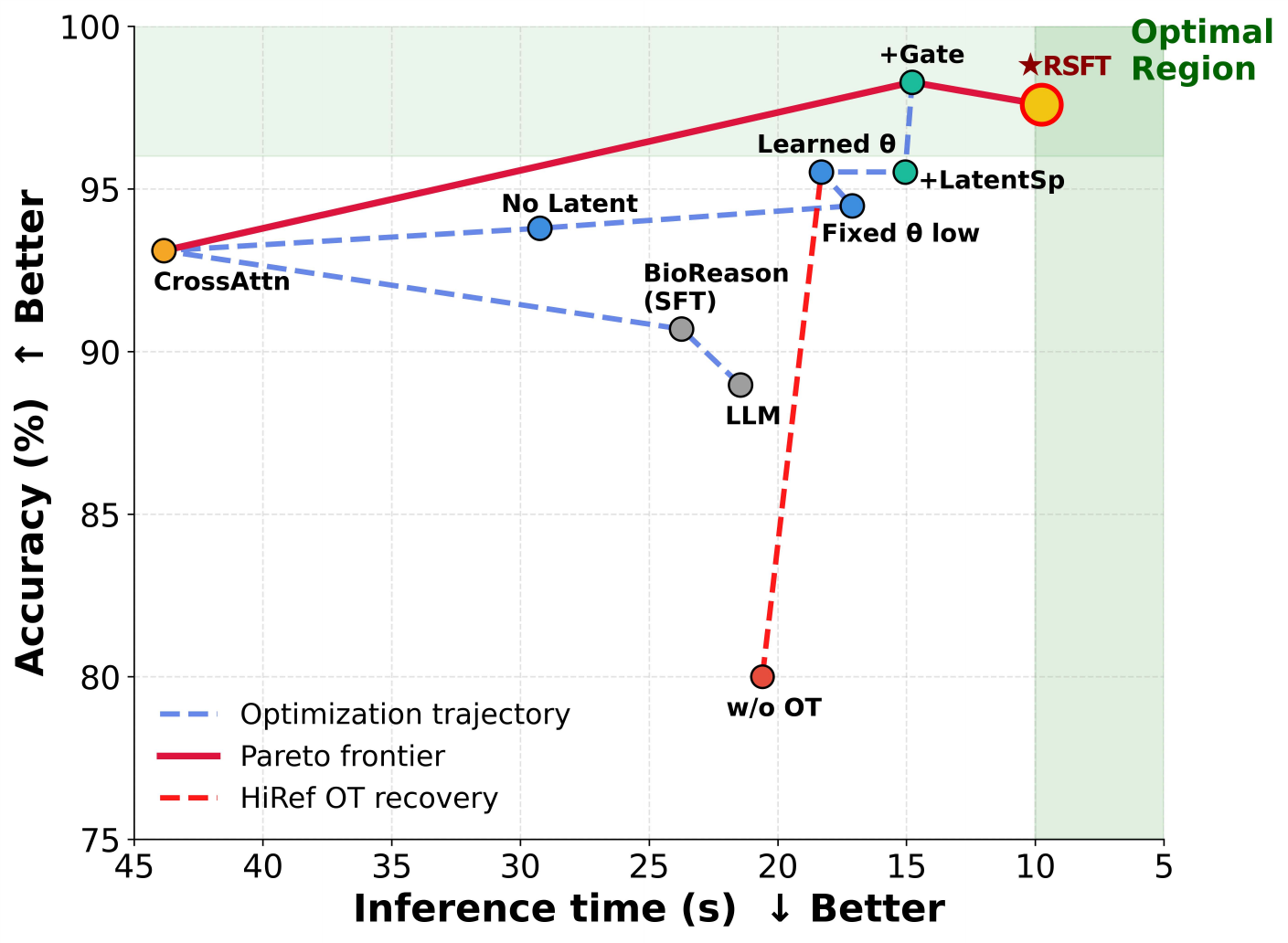}
\caption{Optimization trajectory of {\normalfont\bfseries\itshape GenoMorph} throughout the ablation study in the accuracy--efficiency plane. Each point corresponds to one training stage or architectural variant. The blue dashed curve denotes the optimization trajectory as successive components are incorporated, while the red solid line represents the Pareto frontier. The red dashed segment illustrates the performance recovery obtained by introducing the HiRef OT reward compared with its ablated counterpart. The shaded green region indicates the desirable operating regime with simultaneously high prediction accuracy and low inference time. The final self-adaptive RSFT model achieves the best trade-off between predictive performance and computational efficiency.}
\label{fig:efficiency_plane}
\end{figure}

To quantify the contribution of each component in {\normalfont\bfseries\itshape GenoMorph}, we progressively construct the proposed framework by introducing cross-attention DNA--language fusion, latent-space curriculum learning (LatentSp), adaptive gating, latent reasoning, HiRef OT optimization, and the proposed self-adaptive Rejection Sampling Fine-Tuning (RSFT). Table~\ref{tab:ablation} summarizes the quantitative results, while Figure~\ref{fig:efficiency_plane} visualizes the trade-off between prediction accuracy and inference efficiency throughout the optimization process.

Stage~1 replaces the conventional linear DNA projection with the proposed CrossAttn fusion module. Although this substantially improves the named benchmark accuracy to 93.10\%, inference remains computationally expensive, requiring 43.85~s per sample. Introducing the LatentSp curriculum in Stage~1.5 dramatically accelerates inference, reducing latency by more than 65\% while simultaneously improving predictive performance on both the named and anonymized benchmarks. These results demonstrate that curriculum-guided latent-space optimization stabilizes multimodal reasoning while enabling substantially more efficient genomic representation learning.

Adding the adaptive gating mechanism further strengthens the supervised training stage, achieving 98.28\% accuracy on the named benchmark while maintaining nearly identical inference time. This indicates that selectively activating latent reasoning pathways enables the model to utilize hidden-space computation more effectively than unconditional reasoning prior to reinforcement optimization.

We further investigate different latent reasoning strategies during GRPO optimization. Disabling latent reasoning entirely ($\theta_{\mathrm{low}}=0$) substantially increases inference time while reducing predictive performance. Enabling full latent reasoning using a fixed threshold ($\theta_{\mathrm{low}}=3$) improves computational efficiency but provides only moderate gains in reasoning capability. In contrast, learning the latent activation threshold during training yields the strongest overall robustness, increasing anonymous benchmark accuracy from 74.59\% to 92.07\%. These results demonstrate that adaptive latent activation is considerably more effective than manually selected latent reasoning thresholds.

The contribution of the proposed HiRef OT reward is evaluated through an explicit ablation. Removing this reward decreases the named benchmark accuracy from 95.52\% to 80.00\% and the anonymized benchmark accuracy from 92.07\% to 72.76\%, confirming that hierarchical refinement provides an essential optimization signal for aligning genomic sequence representations with language reasoning. Similarly, removing the LatentSp curriculum substantially reduces both predictive performance and inference efficiency, highlighting its importance for stable latent-space optimization.

Finally, the proposed self-adaptive RSFT produces the strongest overall performance. Compared with the learned-threshold {\normalfont\bfseries\itshape GenoMorph} model, RSFT improves the named benchmark accuracy from 95.52\% to 97.59\% and the anonymized benchmark accuracy from 92.07\% to 94.83\%, while simultaneously reducing inference time by approximately 47\% on the named benchmark (18.31~s to 9.75~s) and 55\% on the anonymized benchmark (23.64~s to 10.73~s). As illustrated in Figure~\ref{fig:efficiency_plane}, the optimization trajectory consistently moves toward the upper-left region of the accuracy--efficiency plane, indicating progressively higher predictive accuracy together with lower computational cost. The Pareto frontier further demonstrates that each major architectural enhancement improves the trade-off between prediction quality and inference efficiency, while the highlighted HiRef OT recovery illustrates the significant performance degradation caused by removing the optimal transport objective. The final self-adaptive RSFT model occupies the most desirable operating point on the Pareto frontier, demonstrating that adaptive latent reasoning successfully optimizes both reasoning quality and computational efficiency without requiring manually specified latent reasoning thresholds.

\begin{table*}[t]
\centering
\caption{Ablation study of {\normalfont\bfseries\itshape GenoMorph}. The table shows the contribution of each architectural component, including Cross-Attention DNA fusion, LatentSp curriculum learning, latent reasoning, adaptive gating, HiRef OT reward, and self-adaptive RSFT.}
\label{tab:ablation}
\resizebox{\textwidth}{!}{
\begin{tabular}{lccccc|ccc|ccc}
\toprule
\multirow{2}{*}{\textbf{Variant}} &
\multirow{2}{*}{\textbf{DNA Fusion}} &
\multirow{2}{*}{\textbf{LatentSp}} &
\multirow{2}{*}{\makecell[c]{\textbf{Latent}\\\textbf{Reasoning}}} &
\multirow{2}{*}{\textbf{Gate}} &
\multirow{2}{*}{\textbf{HiRef OT}} &
\multicolumn{3}{c|}{\textbf{Named Benchmark}} &
\multicolumn{3}{c}{\textbf{Anonymous Benchmark}}\\

\cmidrule(lr){7-9}\cmidrule(lr){10-12}

&
&
&
&
&
&
Acc &
Weighted-F1 &
Time (s) &
Acc &
Weighted-F1 &
Time (s)\\
\midrule

Stage 1: CrossAttn SFT
&
CrossAttn
&
\xmark
&
\xmark
&
\xmark
&
\xmark
&
0.9310
&
0.7573
&
43.85
&
0.5586
&
0.4762
&
26.84
\\

Stage 1.5: +LatentSp
&
CrossAttn
&
\cmark
&
\xmark
&
\xmark
&
\xmark
&
0.9552
&
0.9538
&
15.05
&
0.9034
&
0.8952
&
17.10
\\

Stage 1.5: +Gate
&
CrossAttn
&
\cmark
&
\xmark
&
\cmark
&
\xmark
&
\textbf{0.9828}
&
\textbf{0.9793}
&
14.78
&
0.9069
&
0.9073
&
17.52
\\

GRPO ($\theta_{\mathrm{low}}=0$)
&
CrossAttn
&
\cmark
&
No Latent
&
\cmark
&
\cmark
&
0.9379
&
0.8690
&
29.24
&
0.7518
&
0.7074
&
30.82
\\

GRPO ($\theta_{\mathrm{low}}=3$)
&
CrossAttn
&
\cmark
&
Full Latent
&
\cmark
&
\cmark
&
0.9448
&
0.9067
&
17.11
&
0.7459
&
0.7362
&
18.45
\\

{\normalfont\bfseries\itshape GenoMorph} (Learned $\theta_{\mathrm{low}}$)
&
CrossAttn
&
\cmark
&
Adaptive
&
\cmark
&
\cmark
&
0.9552
&
0.9412
&
18.31
&
0.9207
&
0.9285
&
23.64
\\

\quad w/o HiRef OT
&
CrossAttn
&
\cmark
&
Adaptive
&
\cmark
&
\xmark
&
0.8000
&
0.8250
&
20.60
&
0.7276
&
0.7270
&
19.79
\\

\quad w/o LatentSp
&
CrossAttn
&
\xmark
&
Adaptive
&
\cmark
&
\cmark
&
0.9103
&
0.9135
&
25.40
&
0.7000
&
0.7118
&
31.75
\\

\textbf{\textit{GenoMorph} + Self-Adaptive RSFT}
&
CrossAttn
&
Self
&
Adaptive
&
\xmark
&
\xmark
&
\textbf{0.9759}
&
\textbf{0.9725}
&
\textbf{9.75}
&
\textbf{0.9483}
&
\textbf{0.9465}
&
\textbf{10.73}
\\

\bottomrule
\end{tabular}
}
\end{table*}

\subsection{Qualitative Analysis}

We qualitatively analyze {\normalfont\bfseries\itshape GenoMorph} by examining representative prediction examples together with the remaining failure cases. The case studies demonstrate how adaptive latent reasoning enables the model to integrate genomic sequence information with disease-specific biological knowledge, while the error analysis provides insight into the limitations of the current system.

\subsubsection{Case Studies}

\begin{figure*}[t]
\centering
\includegraphics[width=\textwidth]{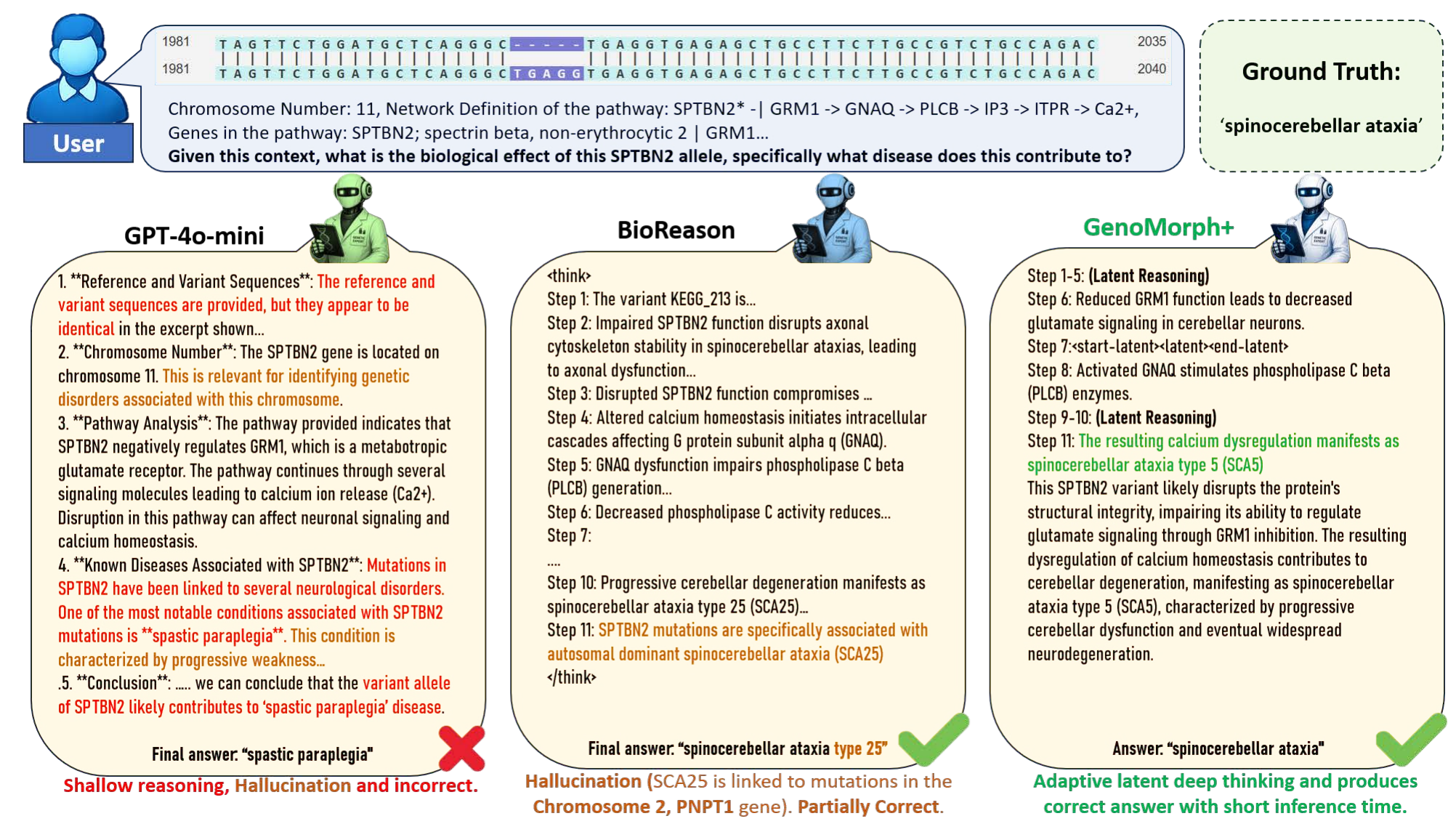}
\caption{Qualitative comparison on the original benchmark. {\normalfont\bfseries\itshape GenoMorph} produces biologically grounded reasoning and correctly predicts the target disease, whereas GPT-4o-mini exhibit wrong reasoning hallucinations leading to incorrect diagnoses and BioReason hallucinates to partially correct diagnoses. }
\label{fig:qualitative_real}
\end{figure*}

\begin{figure*}[t]
\centering
\includegraphics[width=\textwidth]{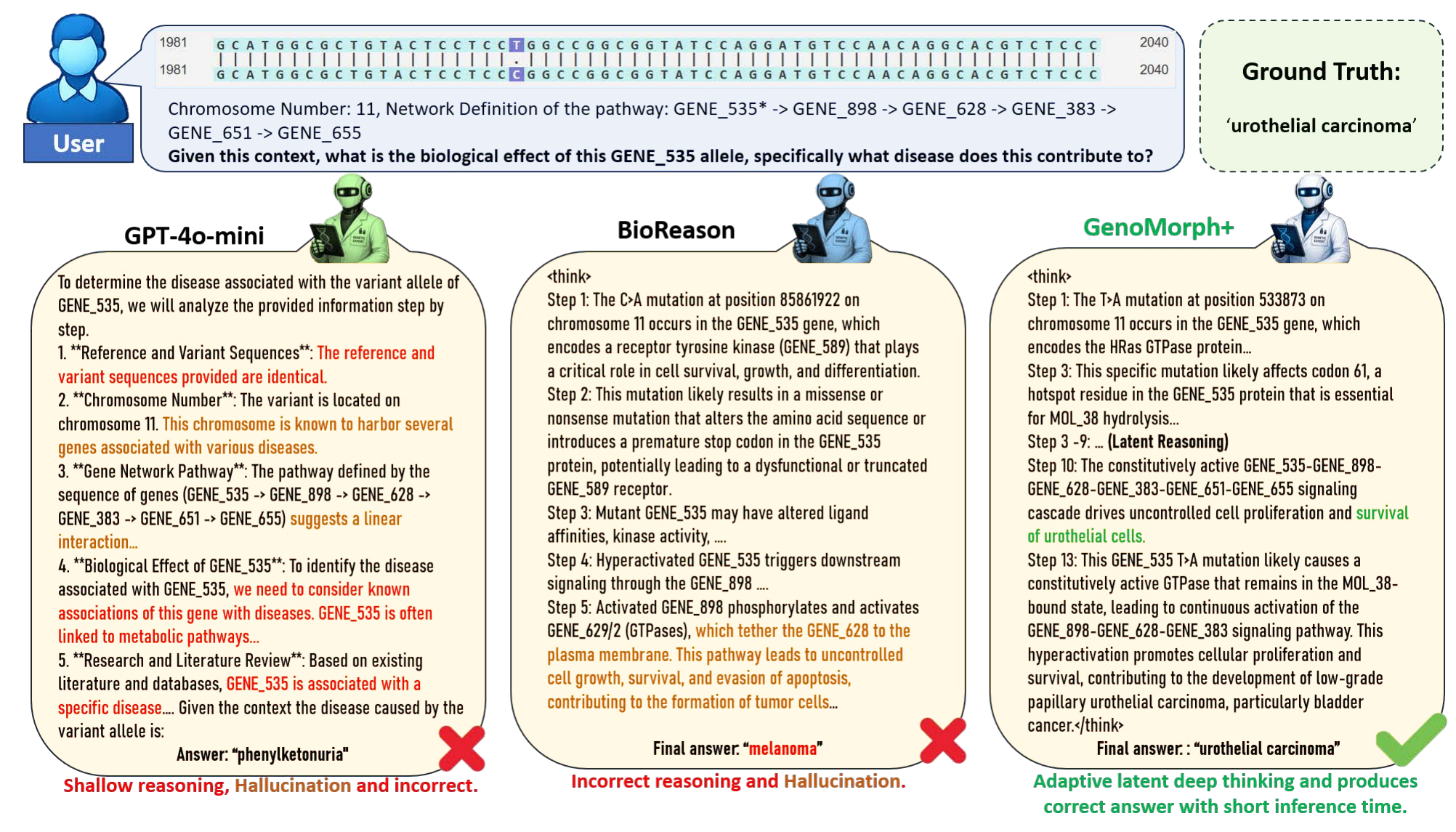}
\caption{Qualitative comparison after gene-name anonymization. Despite the removal of explicit gene identifiers, {\normalfont\bfseries\itshape GenoMorph} correctly infers the disease by reasoning from genomic sequence information, whereas GPT-4o-mini and BioReason fail under the anonymized setting.}
\label{fig:qualitative_anon}
\end{figure*}

Figures~\ref{fig:qualitative_real} and~\ref{fig:qualitative_anon} provide representative examples illustrating the reasoning behavior of {\normalfont\bfseries\itshape GenoMorph} before and after gene-name anonymization. In both cases, the model successfully identifies the correct disease while producing biologically coherent reasoning grounded in the input DNA sequence.

Figure~\ref{fig:qualitative_real} presents an example from the original benchmark containing explicit gene identifiers. GPT-4o-mini produces an incorrect prediction and hallucinates several biological associations that are not supported by the genomic evidence. BioReason partially improves the reasoning process by incorporating DNA sequence information but still relies on inaccurate biological interpretations, leading to an incorrect diagnosis. In contrast, {\normalfont\bfseries\itshape GenoMorph} progressively analyzes the functional role of the reported sequence variants, identifies disrupted biological pathways, and correctly associates these genomic alterations with the target disease. The generated explanation is more structured, biologically consistent, and closely aligned with the known molecular mechanisms.

Figure~\ref{fig:qualitative_anon} evaluates the same reasoning process after replacing all gene names with globally consistent anonymous identifiers. Under this substantially more challenging setting, GPT-4o-mini fails because the removal of explicit biological terminology eliminates the memorized associations on which its prediction depends. BioReason also produces an incorrect diagnosis despite access to DNA representations, indicating limited robustness to identifier anonymization. In contrast, {\normalfont\bfseries\itshape GenoMorph} continues to generate the correct prediction by reasoning directly from the DNA sequence and inferred molecular functions rather than relying on explicit gene names. The generated explanation demonstrates that the model identifies functional genomic patterns and disrupted biological processes even when semantic identifiers are unavailable.

These examples qualitatively support the quantitative results presented in Section~\ref{sec:results}. Together with the anonymization experiments, they demonstrate that {\normalfont\bfseries\itshape GenoMorph} performs biologically grounded genomic reasoning instead of memorizing gene--disease associations. The adaptive latent reasoning framework progressively refines intermediate hypotheses, suppresses unsupported reasoning paths, and produces more faithful explanations with substantially fewer hallucinations than both GPT-4o-mini and BioReason.

\subsubsection{Error Analysis}

To better understand the remaining prediction errors, we manually
analyzed all incorrect predictions produced by the final {\normalfont\bfseries\itshape GenoMorph} model
after self-adaptive RSFT. Among the 13 misclassified samples, the majority
do not correspond to failures in genomic reasoning but instead arise from
differences in disease label granularity.
 
The most common error occurs for Creutzfeldt--Jakob disease (CJD), where
seven samples are predicted as the broader diagnostic category prion
disease. Similarly, one sample belonging to the disease family
sphingolipidoses is predicted as a specific subtype (e.g., Gaucher
disease) rather than the higher-level disease class. In both cases, the
predicted disease is biologically consistent with the underlying genomic
evidence but differs from the benchmark annotation due to hierarchical
disease nomenclature.
 
Table~\ref{tab:error_analysis} summarizes the full breakdown of the 13
misclassified samples. Only six predictions correspond to genuine
classification errors, consisting of two cases of hepatocellular
carcinoma and one case each of pancreatic ductal adenocarcinoma,
colorectal cancer, and non-small cell lung cancer. These samples
represent true capability limitations rather than semantic variations in
disease naming.
 
Overall, approximately 54\% (7/13) of the remaining prediction errors
originate from disease-name granularity rather than incorrect genomic
reasoning, suggesting that the actual biological reasoning capability of
{\normalfont\bfseries\itshape GenoMorph} is stronger than indicated by strict exact-match evaluation.
Future work may further reduce these apparent errors by incorporating
ontology-aware evaluation metrics based on hierarchical disease
relationships.

\begin{table}[t]
\centering
\caption{Manual analysis of the remaining prediction errors produced by {\normalfont\bfseries\itshape GenoMorph} with self-adaptive RSFT.}
\label{tab:error_analysis}
\begin{tabular}{lcl}
\toprule
Disease & Misses & Observation\\
\midrule
Creutzfeldt--Jakob disease & 7 &
\makecell[l]{Predicted broader category\\(Prion disease)}\\

Sphingolipidoses & 1 &
Predicted specific subtype\\

Hepatocellular carcinoma & 2 &
Genuine prediction errors\\

Pancreatic ductal adenocarcinoma & 1 &
Genuine prediction error\\

Colorectal cancer & 1 &
Genuine prediction error\\

Non-small cell lung cancer & 1 &
Genuine prediction error\\
\bottomrule
\end{tabular}
\end{table}
\section{Discussion}

The experimental results consistently demonstrate that explicit integration of genomic sequence representations with adaptive latent reasoning substantially improves genomic disease prediction beyond conventional language-based reasoning. Unlike general-purpose large language models, which rely heavily on memorized associations between gene names and disease concepts, {\normalfont\bfseries\itshape GenoMorph} maintains high predictive performance even after complete anonymization of gene and molecular entity names. The relatively small performance degradation observed under gene-name anonymization indicates that the proposed framework learns biologically meaningful sequence representations rather than depending primarily on lexical memorization. This characteristic is particularly important for practical genomic analysis, where newly discovered genes, anonymized clinical datasets, or previously unseen molecular entities often lack sufficient textual descriptions.

The ablation study further demonstrates that the performance improvements originate from the complementary contributions of multiple architectural components rather than any single modification. Cross-attention fusion effectively aligns DNA sequence embeddings with language representations, while the LatentSp curriculum substantially improves optimization stability and reduces inference latency. Adaptive gating enables selective utilization of hidden-space computation, and the proposed HiRef OT reward provides an effective optimization signal for aligning genomic representations with language reasoning. Collectively, these components progressively move the model toward a superior accuracy--efficiency trade-off, culminating in the proposed self-adaptive RSFT framework, which achieves both the highest predictive performance and the lowest inference latency among all evaluated variants.

An important observation from the qualitative error analysis is that a substantial proportion of the remaining prediction errors arise from differences in disease nomenclature rather than incorrect biological reasoning. More than half of the misclassified samples correspond to biologically consistent predictions that differ from the benchmark annotations because of hierarchical disease relationships, such as predicting the broader category \textit{prion disease} instead of the more specific diagnosis \textit{Creutzfeldt--Jakob disease}. These findings suggest that conventional exact-match evaluation may underestimate the true biological reasoning capability of genomic foundation models. Future genomic reasoning benchmarks could therefore benefit from ontology-aware evaluation protocols that account for hierarchical relationships among disease concepts.

From a broader perspective, {\normalfont\bfseries\itshape GenoMorph} demonstrates that adaptive latent reasoning provides an effective mechanism for jointly optimizing reasoning quality and computational efficiency. Unlike fixed latent reasoning strategies that either over-utilize or under-utilize hidden-space computation, the proposed self-adaptive RSFT dynamically regulates latent reasoning according to task complexity. This adaptive behavior allows computational resources to be allocated only when additional reasoning is required, simultaneously improving prediction accuracy while reducing inference time. These observations suggest that adaptive latent reasoning represents a promising direction for developing scalable genomic foundation models capable of efficient biological reasoning.

Despite these encouraging results, several limitations remain. First, the current study evaluates genomic disease reasoning primarily on a KEGG-derived benchmark, and further validation on larger clinical genomic datasets would strengthen the generalizability of the proposed framework. Second, {\normalfont\bfseries\itshape GenoMorph} currently employs a single genomic encoder, and future work may investigate alternative DNA foundation models jointly optimized sequence encoders, or neural architecture search methods — a paradigm pioneered by Zoph and Le~\cite{nas} and recently specialized to biological foundation models by BioArc~\cite{bioarc} — to automatically discover more effective genomic backbone architectures rather than relying on manual selection. Finally, extending the proposed framework to incorporate complementary molecular modalities, including transcriptomics, epigenomics, proteomics, and regulatory genomic information, may further improve reasoning capability for complex diseases involving multiple interacting biological processes.

\section{Conclusion}

In this work, we presented {\normalfont\bfseries\itshape GenoMorph}, a genomic foundation model that integrates DNA sequence representations with adaptive latent reasoning for genomic disease prediction. Unlike conventional language-based approaches that primarily depend on memorized associations between gene names and biological knowledge, {\normalfont\bfseries\itshape GenoMorph} directly reasons over genomic sequence information through cross-attention DNA--language fusion and a self-adaptive latent reasoning framework. By jointly optimizing genomic representation learning and reasoning efficiency, the proposed framework effectively bridges the gap between sequence-level biological information and language-based disease prediction.

Comprehensive experiments on both the original and gene-name anonymized genomic disease reasoning benchmarks demonstrate that {\normalfont\bfseries\itshape GenoMorph} consistently outperforms existing genomic reasoning approaches. The proposed framework achieves superior predictive performance while exhibiting substantially greater robustness under complete gene-name anonymization, indicating that its predictions are driven primarily by genomic sequence representations rather than lexical memorization. Extensive ablation studies further confirm that CrossAttn fusion, LatentSp curriculum learning, adaptive gating, HiRef OT optimization, and self-adaptive RSFT each contribute to the overall improvements in predictive accuracy and computational efficiency.

Beyond predictive performance, this work demonstrates that adaptive latent reasoning provides an effective mechanism for balancing reasoning quality with inference efficiency. The proposed self-adaptive RSFT dynamically allocates hidden-space computation according to task complexity, allowing {\normalfont\bfseries\itshape GenoMorph} to achieve state-of-the-art performance while simultaneously reducing inference latency. The qualitative analysis further shows that most remaining prediction errors arise from semantic differences in disease nomenclature rather than incorrect biological reasoning, suggesting that the proposed framework captures biologically meaningful genomic relationships.

Overall, {\normalfont\bfseries\itshape GenoMorph} provides a practical and scalable framework for genomic disease reasoning from raw DNA sequences and represents a step toward biologically grounded genomic foundation models. Future work will extend the framework to multi-omics reasoning by integrating transcriptomic, epigenomic, proteomic, and regulatory genomic information, while exploring ontology-aware reasoning and evaluation to better model hierarchical relationships among genes, pathways, and diseases.

\section*{Key points}
 
\begin{itemize}
  \item Genomic disease inference with large language models remains largely dependent on memorized gene--disease associations rather than mechanistic understanding of biological pathways. This shortcut learning limits robustness and generalization when explicit molecular identifiers are unavailable.
  \item We present {\normalfont\bfseries\itshape GenoMorph}, a multimodal genomic reasoning framework that integrates a frozen DNA foundation model with question-conditioned cross-attention fusion, self-adaptive latent reasoning, a residual reasoning gate for genomic evidence reinjection, and rejection sampling fine-tuning regularized by hierarchical optimal transport, shifting disease prediction toward pathway-grounded mechanistic reasoning.
  \item On the original KEGG benchmark, {\normalfont\bfseries\itshape GenoMorph} improves the weighted F1 score from 0.7863 (BioReason baseline) to 0.9412 with GRPO and further to 0.9725 with self-adaptive rejection sampling fine-tuning, while reducing inference time from 23.74\,s to 9.75\,s per sample. On an anonymized benchmark with all gene and molecular identifiers removed, {\normalfont\bfseries\itshape GenoMorph} retains a weighted F1 score of 0.9465, demonstrating that its predictions are driven by genomic sequence evidence rather than lexical memorization. \item {\normalfont\bfseries\itshape GenoMorph} advances genomic disease research by providing an efficient, interpretable framework in which self-adaptive latent reasoning and hierarchical optimal transport jointly improve reasoning robustness and computational efficiency, enabling mechanistic disease reasoning that generalizes beyond memorized gene identities.
\end{itemize}
\section*{Acknowledgments}
The authors gratefully acknowledge the Aryabhatta Supercomputing Centre (ASC) at the Indian Institute of Technology Patna, established under the National Supercomputing Mission (NSM), Government of India, for providing the computational resources utilized in this work.
\section*{Biographical Notes}
\textbf{Tanmoy Kanti Halder} is affiliated with the Department of Computer Science and Engineering, Indian Institute of Technology Patna, and Prasannadeb Women's College. His research interests include multimodal medical and biological reasoning.

\textbf{Akash Ghosh} is a Research Scholar in the Department of Computer Science and Engineering, Indian Institute of Technology Patna, and a former Visiting Researcher at MBZUAI. His research interests include natural language processing, multimodal AI, and AI safety and reasoning.

\textbf{Arijit Roy} is an Assistant Professor in the Department of Computer Science and Engineering, Indian Institute of Technology Patna. His research interests include the Internet of Things, sensor-cloud systems, networking applications and medical large language models.

\textbf{Sriparna Saha} is an Associate Professor in the Department of Computer Science and Engineering, Indian Institute of Technology Patna. Her research interests include artificial intelligence, natural language processing, and biomedical and biological machine learning.

\section*{Author contributions statement}
T.K.H. and A.G. conceived and designed the study. T.K.H., A.G., and A.R. conducted the experiments. T.K.H. and A.G. analysed the results. T.K.H. and A.G. wrote and reviewed the manuscript. S.S. managed the project and revised the final manuscript.

\section*{Conflict of interest}
 
The authors declare that they have no competing interests.

\section*{Data availability}

The code developed in this study is openly accessible on GitHub at
\url{https://github.com/IITP-CSE/GeneticAI}. The original and anonymized KEGG benchmark datasets used for training and evaluation are hosted on Hugging Face and can be accessed at
\url{https://huggingface.co/datasets/iit-patna-cse-ai/kegg-anon-global}.

\section*{Supplementary data}

Supplementary data is available at \emph{Briefings in Bioinformatics} online.

\bibliographystyle{oup-plain}
\bibliography{reference}

\end{document}